\documentclass{article}

\PassOptionsToPackage{numbers, compress}{natbib}
\usepackage{neurips_2024}

\usepackage{amsmath,amsfonts,bm}

\def\eqref#1{equation~\ref{#1}}

\def\1{\bm{1}}

\def\vh{{\bm{h}}}

\DeclareMathAlphabet{\mathsfit}{\encodingdefault}{\sfdefault}{m}{sl}
\SetMathAlphabet{\mathsfit}{bold}{\encodingdefault}{\sfdefault}{bx}{n}

\def\gE{{\mathcal{E}}}

\def\gI{{\mathcal{I}}}

\def\gS{{\mathcal{S}}}

\usepackage{amsmath,amsfonts}
\usepackage{algorithmic}
\usepackage{algorithm}
\usepackage{array}
\usepackage{adjustbox}
\usepackage{multirow}
\usepackage{hyperref}
\usepackage{url}
\usepackage[caption=false,font=normalsize,labelfont=sf,textfont=sf]{subfig}
\usepackage{textcomp}
\usepackage{stfloats}
\usepackage{url}
\usepackage{color}
\usepackage{verbatim}
\usepackage{booktabs}
\usepackage{graphicx}
\usepackage[numbers]{natbib}
\usepackage{comment}

\usepackage[utf8]{inputenc} %
\usepackage[T1]{fontenc}    %
\usepackage{hyperref}       %
\usepackage{url}            %
\usepackage{booktabs}       %
\usepackage{amsfonts}       %
\usepackage{nicefrac}       %
\usepackage{microtype}      %
\usepackage{xcolor}         %
\usepackage{caption}
\usepackage{pdflscape}
\usepackage{xltabular}

\newcommand{\gemma}{Gemma}
\newcommand{\gpt}{ChatGPT}
\newcommand{\gptnew}{GPT4}
\newcommand{\gtr}{GTR}
\newcommand{\llama}{LLaMA}
\newcommand{\opt}{OPT}
\newcommand{\gptneo}{GPT-Neo}
\newcommand{\bloom}{BLOOM}
\newcommand{\mistral}{Mistral}
\newcommand{\qwen}{Qwen}

\newcommand{\phii}{Phi}

\newtheorem{definition}{Definition}

\title{When Text Embedding Meets Large Language Model: A Comprehensive Survey}

\author{%
  Zhijie Nie, Zhangchi Feng, Mingxin Li, Cunwang Zhang, and Richong Zhang\thanks{Corresponding author: zhangrichong@buaa.edu.cn} \\
  School of Computer Science and Engineering,
  Beihang University
  \And
  Yanzhao Zhang, Dingkun Long  \\
}

\begin{document}

\maketitle

\begin{abstract}
Text embedding has become a foundational technology in natural language processing (NLP) during the deep learning era, driving advancements across a wide array of downstream tasks. While many natural language understanding challenges can now be modeled using generative paradigms and leverage the robust generative and comprehension capabilities of large language models (LLMs), numerous practical applications, such as semantic matching, clustering, and information retrieval, continue to rely on text embeddings for their efficiency and effectiveness. Therefore, integrating LLMs with text embeddings has become a major research focus in recent years. In this survey, we categorize the interplay between LLMs and text embeddings into three overarching themes: (1) LLM-augmented text embedding, enhancing traditional embedding methods with LLMs; (2) LLMs as text embedders, adapting their innate capabilities for high-quality embedding; and (3) Text embedding understanding with LLMs, leveraging LLMs to analyze and interpret embeddings. By organizing recent works based on interaction patterns rather than specific downstream applications, we offer a novel and systematic overview of contributions from various research and application domains in the era of LLMs. Furthermore, we highlight the unresolved challenges that persisted in the pre-LLM era with pre-trained language models (PLMs) and explore the emerging obstacles brought forth by LLMs. Building on this analysis, we outline prospective directions for the evolution of text embedding, addressing both theoretical and practical opportunities in the rapidly advancing landscape of NLP.
\end{abstract}

\section{Introduction}
\label{sec:introduction}
Text embedding learning is a fundamental task in Natural Language Processing (NLP), aiming to extract features from text at various levels, including words, sentences, and documents. Formally, given the text $\mathcal{X}$, text embedding learning aims at training a model $f: \mathcal{X} \times \mathrm{\theta} \rightarrow \mathbb{R}^d$ on the dataset $D \subset \mathcal{X}$. The dense vector $h = f_\theta(x)$ output by $f$ is called the embedding of the text $x$, which is expected to contain valid information in the input text and perform well on the downstream tasks.

Advanced large language models (LLMs) have recently demonstrated exceptional generalization capabilities in downstream tasks where generative paradigms are applicable, such as information extraction, text classification, and machine translation. However, not all NLP tasks are suitable for modeling with generative paradigms; tasks such as dense retrieval, semantic text similarity, etc., still require text embedding for computing similarity. The powerful semantic understanding capability of the LLMs reveals the ability to use it to obtain high-quality embeddings. However, like traditional encoder-only pre-trained language models (PLMs), decoder-only LLMs face the anisotropy challenge in their native embedding space \cite{ethayarajh2019contextual}. It is manifested by the high similarity of two token embeddings, which does not reflect their semantic similarity. Higher-level text, such as sentences and documents, suffer from similar problems since they often share embedding space with tokens \cite{li2020sentence}. Fortunately, LLMs have comprehension and generation capabilities far beyond previous PLMs, which opens up new opportunities for text-embedding learning. Specifically, LLMs change the existing landscape in two ways: (1) LLMs can act as data annotators or generators for large-scale, high-quality, and fine-grained text datasets, and (2) LLMs can replace current PLMs as the backbone for generating higher-quality text embeddings. Many works have been devoted to introducing LLMs to obtain high-quality text embedding; however, they focus on a single downstream task, and similar methods have been proposed repeatedly in different research fields. This leads to a lack of comprehensive and objective understanding of the role played by LLMs in text embedding research.

At the same time, the development of LLM has introduced many emerging tasks, some of which are highly relevant to text embedding. This survey introduces two novel downstream tasks with the active community: long context compression and text embedding inversion. Long text compression (ICC) aims to compress long text into a compact number of embeddings while preserving key information. Compared to the context length extension, ICC can improve the decoding efficiency of LLMs and has great potential for application in paradigms such as retrieval-augmented generation (RAG). Embedding inversion aims to recover the original text information from its embedding. With the rapid development of vector databases and commercial embedding services, the study of embedding inversion is significant for protecting privacy and security in embedding. Although the two tasks are developed with different motivations, they share a similar learning framework and use LLM to understand the information in the text embedding. 

\begin{figure*}[t]
\centering
\includegraphics[width=\textwidth]{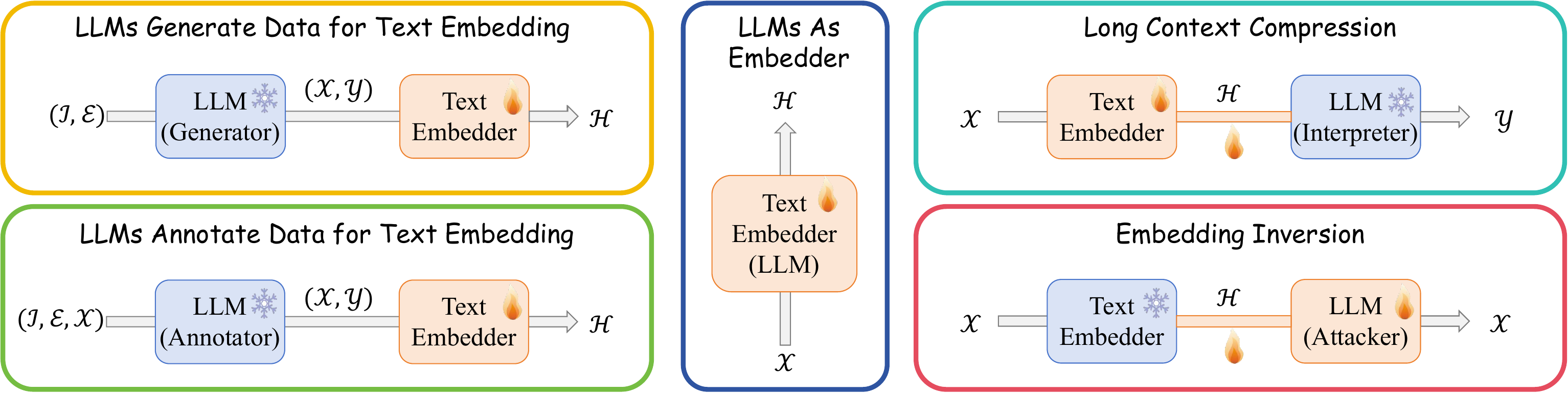}
\caption{An overview of the five relationships between text embedders and LLMs. The correspondence between symbols and meanings is as follows: $\gI$: Instruction, $\gE$: Example, $\mathcal{X}$: Text, $\mathcal{H}$: Text Embedding, $\mathcal{Y}$: Label.}
\label{fig:overview}
\end{figure*}

This survey focuses on deep embedding-learning methods in the LLM era, which include the latest works focused on the traditional downstream tasks related to text embeddings (such as semantic text similarity, information retrieval, and text clustering) and the pioneering work focused on novel downstream tasks (such as long context compression and embedding inversion). Specifically, this survey summarizes the three relationships between LLMs and text embedders for the first time, mainly including (1) LLM-augmented text embedding, (2) LLM as text embedders, and (3) Text embedding understanding with LLMs (Fig. \ref{fig:overview}), while the existing related surveys \cite{cao2024recent,kashyap2023comprehensive,tao2024llms} usually covers only limited works on LLMs embedders. We hope this survey will help researchers in different communities find commonalities in the problem, combining efforts to promote more rapid development of text embedding with the help of LLMs.

The subsequent sections will unfold in the following: Section \ref{sec:priliminary} introduces the developed stage, the training and evaluation tasks for text embedding; Section \ref{sec:llm_augmented} introduces the text embedding methods where LLMs are used for data augmentation and while another model is used to be the text embedder; Section \ref{sec:llm_as_embedder} presents the text embedding methods where LLMs are the backbones of text embedders; Section \ref{sec:embedding_understand} show two novel text embedding-related tasks in the LLM era: long context compression and embedding inversion. Section \ref{sec:challenge} discusses the remaining challenge before the presentation of LLMs and the emerging challenge in the era of LLMs; Section \ref{sec:trends} shares several promising trends and directions in the text embedding field recently.

\section{Priliminary}
\label{sec:priliminary}
\subsection{Brief History of Text Embedding}

\subsubsection{Era of Statistics Machine Learning}
Text embedding is first applied in information retrieval, text mining, and machine translation. In the early days, researchers primarily relied on manually designed features to represent text, which are used to measure the relation between the texts. These methods require domain experts to carefully select and design features, and their effectiveness is constrained by both the quality and quantity of these features. The development of the method is accompanied by a transformation of the output vector's form, which is from bag-of-words models (one-hot vectors) to TF-IDF \cite{ramos2003using} (sparse vectors), then to text embeddings (dense vectors). As machine learning techniques advanced, researchers began exploring methods to learn text embeddings using statistical approaches such as Latent Semantic Analysis (LSA) \cite{landauer1998introduction} for information retrieval and Latent Dirichlet Allocation (LDA) \cite{blei2003latent} for topic mining. Although these methods can automatically learn low-dimensional text embeddings, they still have significant shortcomings. For example, LSA is hard to capture complex semantic and syntactic structures, while LDA is difficult to apply to large-scale datasets.

\subsubsection{Era of Shallow Neural Networks}
With the development of neural machine learning, the NLP community discovered that well-learned embeddings can be transferred to many downstream tasks \cite{collobert2008unified,collobert2011natural,turian2010word}. Therefore, word embedding (distributed representation \cite{hinton1986distributed}) is widely studied as a fundamental technique in NLP instead of a subsidiary to other tasks. We introduce representative text embedding methods in this era from three perspectives: training data, model architecture, and learning paradigm.

\paragraph{Training Data} The training data for word embedding mainly includes two kinds: (1) large-scale text corpus for unsupervised training and (2) high-quality semantic data for secondary supervised training. For example, Wikipedia dump \footnote{\url{https://dumps.wikimedia.org/}}, Gigaword5 \footnote{\url{https://catalog.ldc.upenn.edu/LDC2011T07}} and Common Crawl \footnote{\url{https://commoncrawl.org/}} are usually used for large-scale training corpus. Based on learned word embedding, WordNet~\cite{fellbaum1998wordnet} and Paraphrase Database (PPDB)~\cite{ganitkevitch2013ppdb} are used for semantic knowledge enhancement.

\paragraph{Model Architecture} Though deep neural networks \cite{mikolov2010recurrent, krizhevsky2012imagenet} have shown amazing potential in the same time period, considering the massive size of the corpus, shallow neural network, such as a single-layer Feedforward Neural Network (FNN), is used to maps each word to the embedding.

\paragraph{Learning Paradigm} With the emergence of deep learning techniques, word embedding become a significant breakthrough in the NLP community. (1) Unsupervised paradigm. For word-level, Word2Vec \cite{mikolov2013efficient} proposes Continuous Bag of Words (CBOW) for predicting the center word based on the context and Skip-Gram for predicting the context based on the center word. Word2Vec has two drawbacks: (a) inability to model global statistical Information; (b) inability to handle out-of-vocabulary (OOV) words and the internal structure of words, and these drawbacks are improved by GloVe~\cite{pennington2014glove} and FastText~\cite{bojanowski2017enriching}, respectively. Specifically, GloVe utilizes the global information adequately by pre-constructing the global word co-occurrence matrix, showing that high-quality word embeddings can be obtained without neural networks. FastText represents each word as a collection of n-grams of characters, thus capturing the word's internal structure, which helps to model OOV words. Beyond word-level embedding, Doc2Vec \cite{le2014distributed} proposes Distributed Memory (DM) and Distributed Bag of Words (DBOW) to extend embedding to the document level (e.g., sentences, paragraphs, or entire documents). SIF \cite{arora2017simple} shows that the sentence embedding obtained by weighted averaging of the word embeddings can be regarded as a strong baseline. Besides, removing the projection on the first principal component for each embedding \cite{arora2017simple,mu2018all} will increase the differentiation among these embeddings. (2) Supervised paradigm. Syntactic and semantic knowledge are widely introduced as the supervised signals to improve Word2Vec. For example, DEPS embedding \cite{levy2014dependency} replaces the sliding-window style contexts with dependency-based, where only the modifiers can be the context for the target word. RCM \cite{yu2014improving} and Paragram embedding \cite{wieting2015paraphrase} introduce the semantic relation in Paraphrase Database and WordNet as the constraint to learning the word embedding can reflect these relations.

\subsubsection{Era of Deep Neural Networks} With the development of deep learning, embedding methods with deep neural networks (DNNs) become mainstream and demonstrate remarkable performance on a variety of downstream tasks. Compared to word-level embedding, sentence-level embeddings will be more useful in downstream tasks. Although the simple weighting of word vectors is a strong baseline, it is still worthwhile to explore further how to obtain high-quality sentence embedding on learned word embedding.
\paragraph{Training Data} The labelled datasets for NLP fundamentals and downstream tasks have gradually scaled up, providing the opportunity to learn high-quality text embedding. At the same time, DNNs can be trained on the unlabeled corpus with billions of words with the development of GPUs. Therefore, various datasets for basic NLP research and downstream datasets are widely explored for learning to embed. For example,  BookCorpus~\cite{zhu2015aligning} and the parallel corpus from WMT14~\cite{bojar2014wmt} and WMT15~\cite{bojar2015wmt}; in addition, some popular datasets widely used for embedding learning, such as MS-MARCO \cite{Campos2016MSMA}, SNLI \cite{bowman2015large}, and MNLI \cite{williams2018broad} are proposed.
\paragraph{Model Architecture} Two architectures that can be summarized, which mainly depend on the training task. (1) Encoder-Decoder for generative training tasks. The generative tasks, such as machine translation~\cite{mccann2017learned} and adjacent sentence generation~\cite{kiros2015skip}, are beneficial for generalizing text embeddings. The backbones of decoder can be RNN \cite{cho2014learning}, LSTM \cite{kiros2015skip}, GRU \cite{subramanian2018learning} and Transformer \cite{cer2018universal}, while the backbones of encoder can be CNN \cite{gan2017learning} and DAN \cite{cer2018universal} besides all above models;  (2) Dual-Encoder for discriminative training tasks. The model trained on some discriminative tasks, such as paraphrase matching \cite{wieting2015towards}, natural language inference (NLI) \cite{conneau2017supervised}, and adjacent sentence prediction~\cite{logeswaran2018efficient}, shows the potential of universal sentence embedding in multiple downstream tasks. Except for the standard DNNs \cite{wieting2015towards,conneau2017supervised}, many customized architectures \cite{hu2014convolutional,wang2017bilateral} and pooling strategy \cite{lin2017a} have been explored for training text embeddings.
\paragraph{Learning Paradigm} Three main lines of work for the training paradigm can be summarized: (1) Unsupervised paradigm. The works for unsupervised embedding learning focus on the design of pretext tasks on the unlabeled data. Skip-Thoughts~\cite{kiros2015skip} uses an encoder-decoder architecture that encodes the target sentence and then decodes it to generate the adjacent sentences to learn the embedding. FastSent~\cite{hill2016learning} improves the efficiency of Skip-Thoughts with two efforts: (a) using the sum of word embedding as the sentence embedding for faster encoding speed, and (b) ignoring word order of the adjacent sentences when prediction for fast training speed. Quick-Thoughts~\cite{logeswaran2018efficient} further proposes a discriminative task, training the embedding to select the sentence adjacent to the current sentence in a candidate sentence set. (2) Supervised paradigm. The works under the supervised paradigm focus on finding the specific training task and data on which the embeddings acquired after training can be effectively generalized to other tasks. For example, InferSent~\cite{conneau2017supervised} finds that the embeddings learned on natural language inference (NLI) data have excellent transfer performance in other downstream tasks, while CoVe~\cite{mccann2017learned} finds that the word embedding learned by neural machine translation (NMT) can generalize well to other tasks.  (3) Multi-task paradigm. Some works, such as GenSen \cite{subramanian2018learning} and USE~\cite{cer2018universal}, make use of the successful practical experience in each setting, both of which introduce multiple tasks in the previous two paradigms and obtain the sentence embedding for universal tasks.

\subsubsection{Era of Pre-trained Language Models}\label{sec:cl}

Over the past few years, pre-trained language models (PLMs) with millions of parameters have been proposed first. The ``pretraining then fine-tuning'' paradigms have performed well on various downstream tasks and played an important role in practical on-the-ground applications. However, the word embeddings of the vanilla Transformer~\cite{gao2019representation} and these PLMs~\cite{ethayarajh2019contextual} are proved to be concentrated in high-dimensional conical space, leading to surprisingly high similarities computed for any two words. Therefore, the research centre of text embedding learning shifts to improve the embedding space of PLMs. 

\paragraph{Training Data} Text-embedded data involves multiple downstream applications, and advanced methods often integrate massive amounts of data from various domains in various languages involved in training. Table \ref{tab:training_data} shows non-synthetic datasets often used in existing work. Given the enormous amount of work and the large differences between methods, we will not present the training data used by each method. In addition, current works tend to use zero-shot or few-shot settings to test the generalization ability of text embeddings, so the community has widely accepted the practice of comparing methods with different training data on the same benchmark.

\begin{table*}[ht]
\centering
\caption{Detailed information of available training datasets. The datasets are divided into five major categories, namely information retrieval (IR), question answering (QA), natural language inference (NLI), classification and multi-task (Multi).}
\label{tab:training_data}
\resizebox{\textwidth}{!}{%
\begin{tabular}{l|c|c|c|ccc}
\toprule
\textbf{Dataset} & \textbf{Language} & \textbf{Domain} & \textbf{Task}  &  \multicolumn{3}{c}{\textbf{Text-Text / Text-Label Pair}} \\
\midrule
SNLI~\cite{bowman2015large} & English & Web & NLI & \multicolumn{3}{c}{550,152}  \\
MNLI~\cite{williams2018broad} & English & Web & NLI & \multicolumn{3}{c}{392,702}  \\
AmazonCounter~\cite{oneill2021i} & Multi  & E-commerce  & Classification & \multicolumn{3}{c}{24,000} \\
Emotion~\cite{saravia2018carer} & English & Twitter  & Classification & \multicolumn{3}{c}{16,000} \\
MTOPIntent~\cite{li2021mtop} & Multi & Web  & Classification & \multicolumn{3}{c}{18,800} \\
ToxicConversationsClassification~\footnote{https://www.kaggle.com/competitions/jigsaw-unintended-bias-in-toxicity-classification/overview} & English & Twitter  & Classification & \multicolumn{3}{c}{50,000} \\
TweetSentimentExtraction~\footnote{https://www.kaggle.com/competitions/tweet-sentiment-extraction/overview} & English & Web & Classification & \multicolumn{3}{c}{27,500} \\
\midrule
\textbf{Dataset} & \textbf{Language} & \textbf{Domain} & \textbf{Task} & \textbf{Queries} &  \textbf{Passages} & \textbf{Labels} \\
\midrule
MS MARCO~\cite{Campos2016MSMA} & English & Web & IR & 502,939 & 8,841,823 & 532,761 \\
NFCorpus~\cite{boteva2016} &English & Biomedical & IR & 5,922 & 110,575 & 3,633 \\
SciFact~\cite{wadden2020fact} & English & Scitific  & IR & 809 & 920 & 5,183 \\
BERRI~\cite{asai2023task}  & English & Web & IR & 1,013,774 & 11,187,838 & 1,013,774 \\
DuReader$_\text{retrieval}$~\cite{qiu2022dureader} & Chinese & Web  & IR &97,343 & 8,096,668 & 86,395  \\
Multi-CPR-E-commerce~\cite{long2022multicpr}  & Chinese & E-commerce & IR & 100000 & 1002822 & 100000 \\
Multi-CPR-Video~\cite{long2022multicpr} & Chiense & Video  & IR & 100000 & 1000000 & 100000 \\
Multi-CPR-Biomedical~\cite{long2022multicpr}  & Chinese & Biomedical & IR &  100000 & 959526 & 100000\\
T$^\text{2}$-Ranking~\cite{xie2023t2ranking} & Chinese & Web  & IR & 258,042  &2,303,643 & 1,613,421	\\
mMARCO~\cite{bonifacio2022mmarco}& Multi & Web & IR & 502,939 & 8,841,823 & 532,761 \\
MLDR~\cite{chen2024bge} & Multi & Web   & IR & 41,434 & 493,709 & 41,434 \\
MIRACL~\cite{zhang2023miracl} & Multi & Web  & IR & 40,203	& 90,416,887  & 343,177 \\
Mr.TyDi~\cite{Zhang2021mrtydi} & Multi & Web  & IR & 48,729 & 58,043,326 & 49,127 \\
FEVER~\cite{thorne2018fever} & English & Web  & IR  & 123,142  & 5,416,568 & 140,085 \\
Simclue~\footnote{https://github.com/CLUEbenchmark/SimCLUE} & Chinese & Web & QA & 389,370 & 2,288,523 & 775,593 \\
Natural Questions~\cite{Kwiatkowski2019NaturalQA} & English & Web  & QA & 152,148  & 2,681,468 & 152,148\\
SQuAD~\cite{rajpurkar2016squad} & English & Web  & QA  & 78,713 & 23,215 & 78,713 \\
TriviaQA~\cite{joshi2017triviaqa} & English & Web  & QA & 78,785 & 78,785 & 740K \\
HotpotQA~\cite{yang2018hotpotqa} & English & Web  & QA & 85,000 & 5,233,329 & 170,000  \\
FiQA-2018~\cite{maia2018www18oc}  & English & Finance & QA & 5,500 & 14,166 & 57,638 \\
BioASQ~\cite{tsatsaronis2015overview}  & English & Biomedical & QA & 3,743 & 35,285 & 15,559,157 \\
ArchivalQA~\cite{wang2022archivalqa} & English & News  & QA &853,644 & 853,644  & 483,604 \\
MEDI~\cite{su2023one} & English  & Web  & Multi &  1,240,000 & 1,178,971 & 1,240,000 \\
\bottomrule
\end{tabular}
}
\label{tab:datasets}
\end{table*}

\paragraph{Model Architecture} Most embedding learning works in the era are based on BERT~\cite{kenton2019bert} and RoBERTa~\cite{liu2019roberta}, which consist of multi-layer Transformer Encoders and pre-training on tens of billions of words ($\sim$3.3B words for BERT and $\sim$19B words for RoBERTa). 

\paragraph{Learning Paradigm} (1) Post-hoc paradigm. Pioneering attempts to focus on post-processing methods to improve the quality of the embeddings. For example, adjusting the embedding space to an isotropic space by learning the flow function \cite{li2020sentence} or transforming with whitening matrix \cite{su2021whitening} has proved effective. Besides, standardizing a few undesirable dimensions of the embeddings \cite{timkey2021all} is also an effective post-processing method. 
(2) Unsupervised \& Supervised fine-tuning paradigm. After a brief period of exploration for other loss function \cite{reimers2019sentence}, the supervised and unsupervised paradigms gradually unify to contrastive learning \cite{wu2020clear,karpukhin2020dense,giorgi2021declutr,gao2021simcse,izacard2022unsupervised}. For each sample (``anchor'') in the dataset, Contrastive learning \cite{oord2018representation} focuses on constructing positive and negative examples and optimizing the embedding space to close the anchor-positive distance while large the anchor-negative distance. The difference between the supervised and unsupervised settings is mainly reflected in the construction of positives. In the supervised setting, the anchor-positive pair can be constructed based on the existing labeled dataset, such as query-document pair in the retrieval dataset \cite{karpukhin2020dense} and hypothesis-entailment pair in the NLI dataset \cite{wu2020clear}, etc. In the unsupervised setting, (a) two data-augmented views of the same text \cite{gao2021simcse} or (b) two adjacent texts in the same document \cite{giorgi2021declutr} are regarded as an anchor-positive pair. The augmentation can be literal-level (such as word delete \cite{wu2020clear} and back translation \cite{yan2021consert}, etc) or embedding-level (such as dropout \cite{gao2021simcse}). Negatives can be obtained by random sampling in the dataset, and the hard negative mining methods \cite{zhang2022unsupervised} can be used to discover those challenge negatives with more help for embedding learning. The typical loss function of contrastive learning is InfoNCE \cite{oord2018representation}. Given the anchor sample $x$, the positive example $x^+$ and the negative examples $\{x_j^-\}_{j=1}^N$, denote their $d$-dimensional embedding as $h$, $h^+$ and $\{h_j^-\}_{j=1}^N$, separately, then the InfoNCE Loss is expressed as
\begin{equation}
    \mathcal{L_{\rm cl}} = -\mathbb{E}_{x \sim D} \log \frac{\exp\left(s(h, h^+)\right)}{\exp\left(s(h, h^+)\right) + \sum_j^N \exp\left(s(h, h^-_j)\right)} 
\end{equation}
where $s: \mathbb{R}^d \times \mathbb{R}^d \rightarrow \mathbb{R}$ is the distance function, while $x^+$ and $\{x^-_j\}_{j=1}^N$ is the positive example and negative examples for the text $x$, respectively.  

\paragraph{Technical Improvements}
Improvements based on the contrastive learning have focused on (a) the form of the loss function \cite{zhang2022contrastive,zhou2022debiased,zhuo2023whitenedcse,li2024towards} and (b) better methods for constructing positive and negative samples \cite{yan2021consert,wu2022pcl,wu2022esimcse,zhang2022unsupervised}. For example,
ANCE~\cite{xiong2020approximate} explores a two-stage training method. First, it utilizes non-embedding model methods, such as BM25, to extract a batch of negative samples for model training. Subsequently, it employs the trained embedding model to re-mine higher-quality negative samples for further model training. This method has become a standard practice for training embedding models. Conan-Embedding~\cite{li2024conan} employs a dynamic hard negative mining method to maximize the utilization of high-quality negative examples. However, real-world datasets often suffer from incomplete positive sample labeling due to cost constraints; for most datasets, only around one positive sample is labeled per query. This necessitates careful consideration of the influence of false negatives during negative sample mining. Many researchers suggest that sampling from the middle of the embedding retrieval results or filtering based on similarity scores can effectively enhance model performance~\cite{merrick2024arctic,meng2024sfr}.
(3) Pre-training paradigm.  Contrastive learning is useful on large-scale corpora but takes big performance hits on low data situations \cite{karpukhin2020dense}. This observation shows that the pre-training tasks of PLMs, i.e., Masked Language Modeling (MLM) and Next Sentence Prediction (NSP), etc., do not allow the PLMs to adapt to semantic aggregation quickly and output high-quality text embeddings. Therefore, some work has been devoted to designing better pre-training tasks to improve the performance of contrastive learning in low-data scenarios. Early explorations \cite{lee2019latent,chang2020pre} used only additional incremental pre-training tasks, such as the Inverse Cloze Task (ICT), to improve contrast learning performance. Subsequent improvements involve the modification of model architectures. For example, Condenser \cite{gao2021condenser} restores the masked token using the [CLS]'s hidden states from the last layer and other tokens' hidden states from a shallow layer. Therefore, the newly generated information in the deep layers has to aggregate in the [CLS] to reconstruct the masked information. SEED-Encoder \cite{lu2021less} and RetroMAE \cite{xiao2022retromae} feed the final hidden state of [CLS] into an auxiliary decoder to restore the masked information, encouraging the information aggregation.

\subsection{Large Language Model}
We use the term ``large language models'' or ``LLMs'' to exclude encoder-only PLMs (such as BERT and RoBERTa) and deep neural networks with typically smaller parameter numbers (such as LSTM \citep{hochreiter1997long} and GRU \citep{chung2014empirical}). We refer to the language models whose parameter number is more than 1B as LLMs, which contain large-size encoder-decoder-based PLMs and decoder-only PLMs. We also take into account commercial APIs that are explicitly LLM-backed behind them. Therefore, the LLMs we study can be summarized in Table \ref{tab:llms}.

\begin{table}[ht]
    \centering
    \caption{Representative LLMs contributing to advancements in text embedding.}
    \label{tab:llms}
    \begin{tabular}{c|c}
        \toprule
        {\bf Type} & {\bf Representives} \\
        \midrule
        Encoder-Decoder &  T5 \citep{raffel2020exploring}, FLAN-T5 \citep{chung2024scaling} \\
        \midrule
        \multirow{3}{*}{Decoder-Only} & GPT-Neo \citep{gao2020pile} , BLOOM \citep{le2023bloom}, OPT \citep{zhang2022opt} \\
        & Mistral \citep{jiang2023mistral}, LLaMA Series \citep{touvron2023llama, touvron2023llama2, dubey2024llama} \\
        & Vicuna \citep{lin2023vicuna}, Qwen Series \citep{bai2023qwen, yang2024qwen2, qwen2024qwen2.5} \\
        \midrule
        Commercial API & OpenAI GPT \citep{achiam2023gpt}, Gemini \cite{team2023gemini} \\
        \bottomrule
    \end{tabular}
\end{table}

\subsection{Evaluation Task}
The general trend in text embedding is toward cross-task generalization, and thus, mainstream evaluation protocols tend to evaluate an increasing variety of downstream tasks. Here, we introduce the classical downstream tasks used to evaluate embedding quality. Specifically, we give the definition, common evaluation datasets and evaluation metrics for five tasks, including Semantic Textual Similarity (STS), Information Retrieval (IR), Universal Embedding (UE), Long Context Compression (LCC) and Text Inversion (TI).

\subsubsection{Semantic Textual Similarity}
Semantic Textual Similarity (STS) task \cite{agirre2012semeval} is the task to judge whether the similarity calculated by embeddings matches the ratings from humans. Strictly speaking, STS is not a downstream task with a well-defined application scenario but rather an intrinsic evaluation task purposely presented \cite{cer2017semeval}. The motivation is that the distance between the two texts' embedding should reflect the degree of semantic similarity between these two texts.

\paragraph{Evaluation Dataset} The dataset for evaluating STS can be expressed as $D_{\rm eval}^{\rm STS}=\{(x_i^a, x_i^b, c_i)\}_{i=1}^n$, where $(x_i^a, x_i^b)$ is a text pair and $c_i$ is the average similarity between the two texts. The popular evaluation datasets usually contain seven STS datasets for English setting, which contains STS12-16 \cite{agirre2012semeval,agirre2013semeval,agirre2014semeval,agirre2015semeval,agirre2016semeval}, STS-B \citep{cer2017semeval} and SICK-R \citep{marelli2014sick} in SentEval Benchmark \cite{conneau2018senteval} and STS-17 \cite{cer2017semeval} and STS-22 \cite{chen2022semeval} for multi-lingual setting.
\paragraph{Evaluation Metric} For evaluation, all texts $\{x_i^a\}_{i=1}^n \cup \{x_i^b\}_{i=1}^n$ are encoded by the learned embedder. A similarity function, e.g. dot product or cosine similarity, is introduced to measure the semantic similarity between the text pairs in the embedding space. Common evaluation metrics include the Pearson correlation coefficient and Spearman correlation coefficient. 
\begin{itemize}
    \item {\bf Pearson correlation coefficient} primarily assesses the correlation between the ground-truth $c_i$ and the predicted similarity $c_i^p$, which is expressed as
    \begin{equation}
        r_{\rm pearson} = \frac{\sum_{i=1}^n(c_i-\bar{c}_i)(c_i^p-\bar{c}_i^p)}{\sqrt{\sum_i^n(c_i-\bar{c}_i)^2}\sqrt{\sum_i^n(c_i^p-\bar{c}_i^p)^2}}
    \end{equation}
    where $\bar{c}_i$ and $\bar{c}_i^p$ are the mean value of $\{c_i\}_{i=1}^n$ and $\{c_i^p\}_{i=1}^n$, respectively.
    \item {\bf Spearman correlation coefficient} primarily assesses the correlation between the ranks of $c_i$ and $c_i^p$ in their respective lists, which is expressed as
    \begin{equation}
        r_{\rm spearman} = 1- \frac{6\sum_{i=1}^n (r_i-r_i^p)^2}{n(n^2-1)}
    \end{equation}
    where $r_i$ and $r_i^p$ are the rank of $c_i$ in $\{c_i\}_{i=1}^n$ and $c_i^p$ in $\{c_i^p\}_{i=1}^n$, respectively. 
\end{itemize}
Since different methods lead to large differences in the value range of similarity, and we are concerned with relative rankings rather than absolute values in practice \cite{reimers2016task}. Therefore, the Spearman correlation coefficient is used by default.

\subsubsection{Information Retrieval}
Information retrieval (IR) \footnote{Information retrieval in this survey refers to ad-hoc text retrieval by default.} aims to retrieve the most related texts relevant to the query from a large-size candidate set. Modern IR systems involve multiple stages of recalling or re-ranking from different-scale candidate sets. Embedding-based retrieval methods are referred to as dense retrieval, which is often used as a method of recalling or secondary-filtering, placed after sparse retrieval (e.g., BM25 \cite{robertson2009probabilistic}) and before reranking (e.g., the cross-encoder scheme recommended by BERT ~\cite{kenton2019bert}).

\paragraph{Evaluation Dataset} The dataset for evaluating dense retrieval can be written as $D_{\rm eval}^{\rm IR}=\{q_{i}, d_i^{+}\}_{i=1}^{n} \cup \{d_j^-\}_{j=1}^N$, where $q_{i}$ is the query, $d_i^{+}$ are the related document of $q_{i}$ and $d_j^{-}$ is the unrelated text of any $q_{i}$. In the dataset, $N$ is usually much larger than $n$. The popular benchmarks include BEIR \citep{thakur2021beir} for the English setting, MIRACL \cite{zhang2023miracl} and Mr.TyDi \cite{Zhang2021mrtydi} for the multi-lingual setting.

\paragraph{Evaluation Metric} For efficiency of evaluation, all query $\{q_i\}_{i=1}^n$ and all candidate texts $\{d_i\}_{i=1}^n \cup \{d_j\}_{j=1}^N$ can be encoded as embeddings with the learned embedder in advance. Then, a similarity function is introduced to measure the relevance of each query-candidate embedding pair. For each query, all candidate texts will sorted by the predicted relevance in descending and form an ordered candidate list. The common evaluation metrics include Recall Rate, Accuracy, Mean Reciprocal Rank (MRR), Mean Average Precision (MAP) and Normalized Discounted Cumulative Gain (NDCG).

\begin{itemize}
    \item {\bf Recall Rate} calculates a truncated recall value at the $k$-th position of a sorted candidate list.
    \begin{equation}
        \text{Recall}@k =  \frac{T_{q_i}^{(k)}}{S_{q_i}}
    \end{equation}
    where $S_{q_i}$ denotes the total number of relevant texts for query $q_i$, and $T_{q_i}^{(k)}$ denotes the relevant text number in the top-$k$ position of the sorted candidate list.
    \item {\bf Accuracy} calculates the proportion of queries for which the top-$k$ retrieved texts contain the answers, defined as
    \begin{equation}
        \text{Accuracy}@k = \mathbb{I}(T_{q_i}^{(k)}>0),
    \end{equation}
    where $\mathbb{I}(\cdot)$ is an indicator function.
    \item {\bf MRR} calculates the average reciprocal rank of the first retrieved related text over all queries, which can be denoted as
    \begin{equation}
    \text{MRR} =   \frac{1}{\text{R}_{q_i}^{(1)}}
    \end{equation}
    where $\text{R}_{q_i}^{(1)}$ is the rank of the first relevant text in the ordered candidate list.
    \item {\bf MAP} calculates the mean value of Precision over all queries, which can be expressed as
    \begin{equation}
        \text{MAP} =  \frac{1}{S_{q_i}} \sum_{k =1}^{|S_{q_i}|} \text{Precision}@R_{q_i}^{(k)}
    \end{equation}
    where $\text{R}_{q_i}^{(k)}$ is the rank of the $k$-th relevant text in the sorted candidate list and $\text{Precision}@k = T_{q_i}^{(k)} / k$.
    \item {\bf NDCG} considers the rank of the relevant text and suggests situating the more pertinent text at the more superior position. DCG for each $q_i$ should be calculated first, which can be written as
    \begin{equation}
        \text{DCG}_{q_i}@k = \sum_{i=1}^{k}\frac{2^{r_i}-1}{\log_2(i+1)}
    \end{equation}
    where $r_i$ is the graded relevance score for the $i$-th retrieved text in the candidate text list. 
    Then $\text{NDCG}@k$ can be calculated by aggregating the normalized DCG values at a specific rank position:
    \begin{equation}
        \text{NDCG}@k =  \frac{\text{DCG}_q@k}{\text{IDCG}_q@k}
    \end{equation}
    where IDCG@$k$ denote ideal DCG (the DCG value when assuming that the retrieved results are ordered optimally).
\end{itemize}

\subsubsection{Universal Embedding}
While tasks such as classification, rerank, summarization, etc., can be solved using the LLMs' generative paradigm, the state-of-the-art methods require complex handwritten prompts, many in-context examples and long output for chain-of-thought. Using LLM embedding to solve these tasks is still an efficient and practical approach \cite{liu2024llmembed}. Therefore, training an embedder to get a more general embedding for all tasks is also a hot research topic.

\paragraph{Evaluation Dataset} The MTEB benchmark \cite{muennighoff2023mteb} contains 56 datasets in 7 different types of downstream tasks, including classification, clustering, pair classification, reranking, retrieval, and Semantic Textual Similarity (STS), to assess the degree of generalization of text embedding. Until October 2024, MTEB has the community variants in Chinese \cite{xiao2023c}, French \cite{ciancone2024mtebfrench}, Polish \cite{poswiata2024pl}, Scandinavian \cite{enevoldsen2024scandinavian}, Russian \cite{snegirev2024russian}, Arabic \cite{bhatia2024swan} and Persian \cite{zinvandi2025famteb}, and an official multilingual version, i.e. MMTEB \footnote{\url{https://github.com/embeddings-benchmark/mteb/blob/main/docs/mmteb}}, during developing.

\paragraph{Evaluation Metric} MTEB has developed a uniform evaluation protocol for each type of task for a fair comparison. Please refer to their original paper \cite{muennighoff2023mteb} and official GitHub repository \footnote{\url{https://github.com/embeddings-benchmark/mteb}} for the details of the evaluation metric.

\subsection{Emerging Related Tasks}

\subsubsection{Long Context Compression}
Long Context Compression (ICC) aims to compress long context into text embedding or token sequence for LLMs without sacrificing essential information, accelerating inference while maintaining generation results. LLMs are expected to be interpreters that understand original context information from compressed embedding or token sequence and output content consistent with what it would output when there is no compression.

\paragraph{Evaluation Dataset} The evaluation for LCC usually uses the dataset for Question Answering (QA) or Retrieval Augmented Generation (RAG). The datasets are formalized as $D_{\rm LCC}^{\rm eval} = \{(c_{i}, q_{i}, a_{i})\}_{i=1}^{n}$ where $c_{i}$ represents the $L_{c_i}$-token context, $q_{i}$ represents the question, and $a_{i} = \{t_i^{(j)}\}_{j=1}^{L_{a_i}}$ represents the $L_{a_i}$-token answer.

\paragraph{Evaluation Metric} The embedder for LCC takes the L-length long context $c_i$ as input and outputs a compressed embedding matrix $c_i' \in \mathbb{R}^{d \times l_i}$. The evaluation for LCC can be divided into two aspects: (1) compression efficiency and (2) generation consistency. Specifically, context compression rate~\cite{jiang2023llmlingua} and average inference time~\cite{cheng2024xrag} are proposed to evaluate the compression efficiency, while Perplexity~\cite{ge2023context}, Exact Match~\cite{cheng2024xrag}, and ROUGE-L~\cite{cheng2024xrag} are suggested for the evaluation of generation consistency.
\begin{itemize}
    \item {\bf Context Compression Rate} measures the ratio of the compressed context length to the original context length, which is denoted as
    \begin{equation}
        \text{CompressionRate} =  \frac{l_i}{L_{c_i}}
    \end{equation}
    \item {\bf Perplexity} measures how well the model predicts the next word in the answer, indicating its confidence and fluency:
    \begin{equation}
        \text{Perplexity}= \prod_{i=1}^{L_{a_i}}\frac {1} {p(t_i^{(j)}|t_i^{(1)},...,t_i^{(j-1)})})^{\frac{1}{L_{a_i}}}
    \end{equation}
    \item {\bf Exact Match (EM)} checks if the generated answer exactly matches the reference answer, rewarding only perfect matches:
    \begin{equation}\label{eqn:em}
        \text{EM} =  \frac{\sum_{i=1}^{\operatorname{min}\{L_{a_i}, L_{\hat{a}_i}\}} \mathbb{I}(\hat{t}_{a_i}^{(j)} = t_i^{(j)})}{L_{a_i}}
    \end{equation}
    where $\hat{t}_{a_i}^{(j)}$ is the $j$-th token in the generated answer $\hat{a}_{i}$ with the compressed context and $L_{\hat{a}}$ is the length of $\hat{a}_{i}$.
    \item {\bf ROUGE-L} evaluates the overlap between the generated answer and the reference answer by focusing on the longest common subsequence, allowing for partial correctness:
    \begin{equation}\label{eqn:rouge-l}
        \text{ROUGE-L}_{\text{R}} = \frac{\operatorname{LCS}(a_i, \hat{a}_i)}{L_{a_i}}
    \end{equation}
    \begin{equation}
        \text{ROUGE-L}_{\text{P}} = \frac{\operatorname{LCS}(a_i, \hat{a}_i)}{L_{\hat{a}_i}}
    \end{equation}
    \begin{equation}
        \text{ROUGE-L} = \frac{(1 + \beta^2) \cdot \text{ROUGE-L}_{\text{P}} \cdot \text{ROUGE-L}_{\text{R}}}{\beta^2 \cdot \text{ROUGE-L}_{\text{P}} + \text{ROUGE-L}_{\text{R}}}
    \end{equation}
\end{itemize}
where $\beta$ is a hyper-parameter that controls the relative importance of precision and recall. When $\beta$ is 1, the ROUGE-L score is simply the harmonic mean of precision and recall. The greater the $\beta$, the heavier the recall's weight.

\subsubsection{Embedding Inversion}
Embedding Inversion (IC) predicts part or all text information based on the embeddings. It is generally regarded as an attacker, with targets including attribute information in the input text, the entire input text, or the text embedding model itself.
\paragraph{Evaluation Dataset} For different levels of leak information, the datasets can be word-level or sentence-level. For attribute inference attack, the dataset has the form of (Input, Words), where the "Input" is text and "words" represents important information in the input text, for example, name and ID number. To reconstruct all the input text, the dataset just contains input text. Note that text embedding will not be contained in the dataset directly; it can be generated by the victim embedding model with the text in the dataset.

\paragraph{Evaluation Metric} The main metrics for evaluating embedding inversion include BLEU, ROUGE, Exact-Match, cosine similarity, and Token F1. Please refer to Equation \ref{eqn:em} and \ref{eqn:rouge-l} for Exact-Match and ROUGE-L, separately.

\begin{itemize}
    \item {\bf BLEU} measures n-gram similarity between reconstructed text and original input text. We can compute it in the following:
    \begin{equation}
        \text{BLEU} = \text{BP}\times \exp{\sum_{n=1}^{N}w_nlog(p_n)}
    \end{equation}
    
    \item {\bf Cosine Similarity} evaluates the similarity between the embeddings of the reconstructed text and the original input text in the embedding space. It is defined as:
    \begin{equation}
    \text{Cosine Similarity} = \frac{\mathbf{a}\cdot\mathbf{b}}{\|\mathbf{a}\|\|\mathbf{b}\|}
    \end{equation}
    where $\mathbf{a}$ and $\mathbf{b}$ are the embeddings of the reconstructed and original texts, respectively.
    
    \item {\bf Token F1} is a word-level metric that computes the F1 score between tokens of the predicted text and the input text. It is defined as the harmonic mean of token-level precision and recall:
    \begin{equation}
    \text{Token F1} = 2\times \frac{\text{Precision}\times \text{Recall}}{\text{Precision} + \text{Recall}}
    \end{equation}
where BP is the brevity penalty, $w_n$ is the weight for each n-gram precision, and $p_n$ is the modified precision for n-th n-gram.
\end{itemize}

\begin{table*}
    \centering
    \caption{The Overview of LLM-augmented Text Embedding. I, Q, D$^+$, D$^-$, L denote instruction, query, positive document, hard negative document, and pair similarity separately.}
    \resizebox{\textwidth}{!}{\begin{tabular}{llllllll}
    \toprule
    \multirow{2}{*}{\bf Method} & \multicolumn{2}{c}{\bf Model} & \multicolumn{3}{c}{\bf Generated Training Data} & \multicolumn{2}{c}{\bf Evaluation} \\
    \cmidrule(r){2-3} \cmidrule(r){4-6} \cmidrule(r){7-8}
     & LLM & Encoder & Scale & Form & Open-Source & Protocol & Dataset \\
    \midrule
    SKICSE \cite{ou2024skicse} & LLaMA2 & BERT & 1M, 276K & (Q, D$^+$) & $\times$ & ZS & STS \\
    DenoSent \cite{wang2024denosent} & ChatGPT & BERT, RoBERTa & 1M & (Q, D$^+$) & $\surd$ & ZS & STS \\
    GenSE \cite{chen2022generate} & T5$^*$ & T5 & 61M & (Q, D$^+$, D$^-$) & $\surd$ & FS & STS \\
    SynCSE \cite{zhang2023contrastive} & ChatGPT, GPT4 & RoBERTa & 276K & (Q, D$^+$, D$^-$) & $\surd$ & ZS & STS \\
    MultiCSR \cite{wang2023semantic} & ChatGPT, FLAN-T5$^*$ & BERT, RoBERTa & 1M, 276K & (Q, D$^+$, D$^-$) & $\surd$ & ZS & STS \\
    AoE \cite{li2024aoe} & ChatGPT, LLaMA2, ChatGLM & BERT & 6K & (Q, D$^+$, D$^-$) & $\surd$ & ZS & STS \\
    Work from \cite{sato2024improving} & LLaMA2 & LLaMA2 & 256K & (Q, D$^+$, D$^-$) & $\surd$ & ZS & STS \\
    SumCSE \cite{thirukovalluru2024sumcse} & Vicuna & RoBERTa & 276K & (Q, D$^+$, D$^-$) & $\surd$ & ZS & STS \\
    AdaptCL \cite{xu2024adaptive} & WizardLM$^*$ & BERT, RoBERTa, T5 & 60K & (Q, D$^+$, D$^-$) & $\surd$ & ZS & STS \\
    CLHAIF \cite{cheng2023improving} & GPT3 & BERT, RoBERTa & 276K & (Q, D$^+$, L) & $\surd$ & ZS & STS \\
    CLAIF \cite{cheng2023improving} & GPT3 & BERT, RoBERTa & 113K, 1.2M & (Q, D, L) & $\surd$ & ZS & STS \\
    NGCSE \cite{li2024narrowing} & ChatGPT & BERT, RoBERTa & 60K & (Q, D, L) & $\surd$ & FS & STS \\
    \midrule
    Promptagator \cite{dai2022promptagator} & FLAN-T5 & T5 & 8M & (Q, D$^+$) & $\times$ & ZS / FS & IR \\
    Work from \cite{ma2023pre} & Alpaca, tk-Instruct & BERT & 3.2M & (Q, D$^+$)& $\times$ & ZS / FT & IR \\
    SPTAR \cite{peng2025soft} & LLaMA$^*$, Vicuna$^*$ & LLaMA, Vicuna & 100K & (Q, D$^+$)& $\surd$ & FT & IR \\
    InPars \cite{bonifacio2022inpars} & \gpt & monoT5 & 10K & (Q, D$^+$, D$^-$) & $\surd$ & ZS / FS & IR \\
    InPars-v2 \cite{jeronymo2023inpars} & GPT-J & monoT5 & 180K & (Q, D$^+$, D$^-$) & $\surd$ & ZS / FS & IR \\
    NV-Retriever \cite{moreira2024nv} & E5$_{\rm mistral}$ & Mistral & 956K & (Q, D$^+$, D$^-$) & $\times$ & ZS / FT & IR \\
    I3 \cite{pan2024i3} & \gpt & COCO-DR(BERT) & 140K & (I, Q, D$^+$, D$^-$) & $\times$ & ZS & IR \\
    Promptriever \cite{weller2025promptriever} & LLaMA3, GPT4o & LLaMA2 & 491K & (I, D, D$^+$, D$^-$) & $\surd$ & ZS / FT & IR \\
    \midrule
    Gecko \cite{lee2024gecko} & - & 1.2B LLM & 6.6M & (I, Q, D$^+$, D$^-$) & $\times$ & ZS & Uni. \\
    E5$_{\rm mistral}$ \cite{wang2024improving} & \gpt, \gptnew & Mistral & 500K & (I, Q, D$^+$, D$^-$) & $\times$ & ZS & Uni. \\
    FollowIR \cite{weller2024followir} & \gpt & Mistral & 1.8K & (I, Q, D$^+$, D$^-$) & $\surd$ & ZS & Uni. \\
    \bottomrule
    \end{tabular}}
    \label{tab:llm_augmented}
\end{table*}

\section{LLM-Augmented Text Embedding}
\label{sec:llm_augmented}

One approach to adopting LLMs for text embeddings is through knowledge distillation. Specifically, LLMs can be used to generate training data for the embedding model in two ways: 1. Directly synthesizing training data (\S~\ref{sec:03_data_synthesis}); 2. Providing supervision signals for existing data (\S~\ref{sec:03_data_annotation}). Please refer to Table~\ref{tab:llm_augmented} for the methods presented in this survey.

\subsection{Data Synthesis with LLMs}
\label{sec:03_data_synthesis}

Current text embedding models are typically trained using contrastive learning~\cite{oord2018representation}, where the training data commonly consist of three components: anchors, positive samples, and negative samples. The specific definitions of these components vary across tasks. For example: In Information Retrieval (IR), the anchor is the query, while the positive and negative samples are documents related and unrelated to the anchor, respectively. In Semantic Textual Similarity (STS) tasks, the anchor is a text, with positive and negative samples being semantically similar and dissimilar texts. Additionally, recent studies~\cite{su2023one,weller2025promptriever} have begun integrating instruction-following capabilities into text embedding models, resulting in a need for instructions in the training data. In this section, we provide an overview of how LLMs can be used to synthesize the different components of training data.

\subsubsection{Instructions}

Existing instructions in training data are typically constructed based on datasets \cite{su2023one}. This involves manually collecting multi-task datasets and applying templates to generate different instructions for each dataset~\cite{su2023one}. However, this approach lacks diversity, limiting the full potential of the model's instruction-following capabilities. 
Therefore, a promising solution is to generate diverse instructions by using LLMs.
One approach leverages LLMs to generate various instructions by varying the given conditions: I3~\cite{pan2024i3} generates diverse instructions by manually setting different conditions, including topics, organizational formats of the retrieved text, and definitions of relevance; E5$_\mathrm{mistral}$~\cite{wang2024improving} categorizes tasks into different types (e.g., symmetric and asymmetric retrieval tasks) and designs specific prompts for each category to generate instructions. 
Another approach leverages LLMs to generate instance-level instructions for each document, further enhancing the diversity of the instructions: Gecko~\cite{lee2024gecko} prompts LLMs to generate instructions by conditioning on different documents; Promptriever~\cite{weller2025promptriever} generates instructions that describe the relationship between the given queries and documents, and enhance instruction diversity by setting various conditions such as length and style of the generated instructions. 

\subsubsection{Positive Samples}
\label{sec:03_positive_samples}

Based on the formal similarity between positive samples and anchors, positive samples can generally be classified into the following two categories.

\paragraph{Symmetric Positive Samples} A classic task where anchors and positive samples are symmetric is STS. In this area, numerous studies have proposed various methods to generate symmetric positive samples. Based on the analysis in S-BERT~\cite{reimers2019sentence}, text embedding models for STS are often trained using training data from the Natural Language Inference (NLI) task. Following this convention, a line of research explores how to use LLMs to generate NLI-style training data: NGCSE~\cite{li2024narrowing} directly utilize the in-context learning capabilities of LLMs by using NLI-formatted training data as demonstrations to guide the generation of entailment sentences as positive samples; 
GenSE~\cite{chen2022generate} and MultiCSR~\cite{wang2023semantic} first fine-tune LLMs with NLI training data, enabling them to generate NLI-formatted positive samples. Apart from supervised fine-tuning, AdaptCL~\cite{xu2024adaptive} employ a reinforcement learning approach to iteratively update both the data generation model and the text embedding model, achieving improved performance. In addition, many works focus on exploring generating other types of semantically similar texts as positive samples: SynCSE
~\cite{zhang2023contrastive} and AoE~\cite{li2024aoe} generate diverse semantically similar positive samples by setting different conditions (e.g., genre, topic) and leveraging the in-context learning capabilities of LLMs; SKICSE~\cite{ou2024skicse} constructs positive samples by extracting additional information about the anchor using the inherent knowledge of LLMs. SumCSE~\cite{thirukovalluru2024sumcse} generates positive samples by having LLMs summarize the anchor; CLAIF~\cite{cheng2023improving} creates positive samples by masking tokens in the anchor and completing the missing parts using LLMs; E5$_\mathrm{mistral}$~\cite{wang2024improving} further subdivides symmetric retrieval tasks into STS tasks and bitext retrieval tasks, designing specific prompts to generate different types of positive samples for each.

\paragraph{Asymmetric Positive Samples} In asymmetric cases, the anchor and positive sample are often referred to as the query and document, respectively. In various studies, either the query or the document can be the target for generation by LLMs. Many works~\cite{dai2022promptagator,bonifacio2022inpars,jeronymo2023inpars,ma2023pre, peng2025soft,lee2024gecko} use the vast amount of documents available on the internet to generate queries: Some of them~\cite{dai2022promptagator,bonifacio2022inpars,jeronymo2023inpars,ma2023pre, lee2024gecko} directly leverage the in-context learning capability of LLMs, constructing prompts to facilitate query generation. Notably, InPars~\cite{bonifacio2022inpars,jeronymo2023inpars} introduces a ``Guided by Bad Questions (GBQ)'' template, which uses randomly sampled pseudo-queries to improve the quality of generated queries; Moreover, SPTAR~\cite{peng2025soft} employs soft prompts to fine-tune LLMs, enabling them to generate queries based on documents. In addition to generating queries from documents, there are other generation methods: I3~\cite{pan2024i3} and E5$_\mathrm{mistral}$~\cite{wang2024improving} generate both queries and documents based on their self-generated instructions; FollowIR~\cite{weller2024followir} and promptriever~\cite{weller2025promptriever} guides document generation using queries and the generated instructions.

\subsubsection{Negative Samples}
\label{sec:03_negative_samples}

Negative samples are crucial in contrastive learning, and research~\cite{robinson2021contrastive} has shown that hard negatives, as opposed to randomly sampled negatives from training batches, can significantly improve model performance. Consequently, many studies have explored methods for generating hard negative samples using LLMs. When generating hard negatives, methods similar to those for generating positive samples (as discussed in \S~\ref{sec:03_positive_samples}) are often employed~\cite{chen2022generate,wang2023semantic,sato2024improving,xu2024adaptive,li2024narrowing,zhang2023contrastive,li2024aoe,thirukovalluru2024sumcse,weller2025promptriever}. The generated hard negatives can generally be divided into two categories. 1. NLI-based negatives~\cite{chen2022generate,wang2023semantic,sato2024improving,xu2024adaptive,li2024narrowing}: Texts with a logical relationship of contradiction are generated as hard negatives; 2. Contextually irrelevant negatives~\cite{zhang2023contrastive,li2024aoe,thirukovalluru2024sumcse,weller2025promptriever}: Texts are generated using specific definitions of irrelevance as hard negatives. Notably, Promptriever~\cite{weller2025promptriever} introduces a unique type of hard negative to enhance the model's instruction-following capability. These hard negatives are positive samples relative to the query itself but become negative samples when the query is combined with the instruction.

\subsection{Data Annotation with LLMs}
\label{sec:03_data_annotation}

In many scenarios, lots of existing data are available, where leveraging LLMs to provide supervision signals for data annotation is a more effective and efficient approach. In this section, we introduce the use of supervision signals from LLMs in the following scenarios: Mining training data from existing corpus; Filtering incorrect supervision signals in existing training data; Providing supervision signals for clustering algorithms.

\subsubsection{Training Data Supervision}

Compared to directly generating text from LLMs to construct training data, mining training data from large corpus is more effective in meeting the demand for diversity. As introduced in \S~\ref{sec:03_negative_samples}, hard negatives are critical for text embedding models. Therefore, some studies leverage LLMs for hard negative mining: NV-Retriever~\cite{moreira2024nv} uses LLMs as embedding model and employs proposed ``positive-aware mining methods'' to identify hard negatives; Gecko~\cite{lee2024gecko} first retrieves relevant texts using a embedding model, then estimates the similarity between texts using LLMs to mine hard negatives. In addition to traditional training data composed of positive and negative samples, the application of LLMs enables the construction of training data with finer-grained similarity patterns: CLAIF and CLHALF~\cite{cheng2023improving} use LLMs to assign similarity scores to generated and existing text pairs, then train models using Mean Squared Error (MSE) loss and a proposed soft InfoNCE loss; NGCSE~\cite{li2024narrowing} employs LLMs to construct text pairs with similarity levels ranging from high to low, which equals to labeling text pairs with similarity scores. These pairs are then used for training with a proposed Hierarchical Triplet loss.

\subsubsection{Training Data Filtering}

Existing training data and newly constructed training data from LLMs may both contain incorrect supervision signals. Therefore, filtering or even correcting erroneous supervision signals is an important research direction. This section first introduces methods for filtering data constructed by LLMs, followed by methods using LLMs for data filtering.

\paragraph{Filtering Training Data From LLMs} Erroneous supervision signals can be categorized into two types: false negatives and false positives. Several studies focus on methods for filtering false negatives: MultiCSR~\cite{wang2023semantic} filters false negatives by using a high-performing embedding model to provide text similarity scores; Promptriever~\cite{weller2025promptriever} employs a cross-encoder to compute text similarity for filtering false negatives. Other studies focus on excluding false positives that are not in the Top-K retrieved results: Promptagator~\cite{dai2022promptagator} trains a retriever on generated data; InPars-v2~\cite{jeronymo2023inpars} trains a retriever using MS MARCO data; SPTAR~\cite{peng2025soft} directly uses BM25 as the retriever. In addition, AdaptCL~\cite{xu2024adaptive} utilizes an NLI classification model and FollowIR~\cite{weller2024followir} utilizes an top-tier embedding model to filter both false negatives and false positives simultaneously. 

\paragraph{LLMs as Training Data Filter} Some studies use LLMs to filter false positives in generated data: InPars~\cite{bonifacio2022inpars} filters false positives using perplexity as a metric during text generation; I3\cite{pan2024i3} designs prompts to filter out queries that do not meet specified conditions. Additionally, some studies employ LLMs to filter false positives and false negatives in generated data simultaneously: GenSE fine-tunes LLMs to identify incorrect entailment and contradiction; MultiCSR~\cite{wang2023semantic} prompts LLMs to filter out instances that fail to meet the conditions for entailment and contradiction.

\subsubsection{Clustering Supervision} 

Embedding models are often applied in clustering algorithms, and many recent studies leverage LLMs to provide supervision signals for clustering processes, enhancing task performance. KeyEvents~\cite{nakshatri2023using} uses LLMs to assist in Key Event Discovery by aggregating clustering results and summarizing cluster centers to identify key events. Two studies employ LLMs to support Generalized Category Discovery: ALUP~\cite{liang2024actively} introduces an uncertainty metric to identify error-prone samples in the clustering results and uses LLMs to reassign these samples to different cluster centers. LOOP~\cite{an2023generalized} identifies boundary samples as error-prone and employs LLMs to determine which category boundary these samples are closer to. In LOOP, LLMs are also used to extract the semantic meaning of each cluster. IDAS~\cite{de2023idas} assists in the Intent Discovery task by using LLMs to label each sample, improving clustering performance. ClusterLLM~\cite{zhang2023clusterllm} enhances clustering algorithms by using LLMs to generate triplet constraints on an initial clustering result, applying hierarchical clustering to derive the final result, and validating the correctness of cluster merges during the process with LLMs. Another study~\cite{viswanathan2023large} integrates LLMs throughout the entire clustering process: Pre-clustering, LLMs extract sample keywords to improve feature representation; During clustering, LLMs provide pairwise constraints to enhance clustering quality; Post-clustering, LLMs reassign error-prone samples to the correct clusters.

\section{LLMs as Text Embedder}
\label{sec:llm_as_embedder}

LLMs commonly adopt encoder-decoder architecture, e.g., T5~\cite{raffel2020exploring}, or decoder-only architecture, e.g., GPT-3~\cite{brown2020language}. After going through pre-training, instruction tuning, and alignment with human preferences~\cite {ouyang2022training}, LLMs can perform a wide range of tasks. However, these steps do not support obtaining high-quality embeddings from LLMs. To address this issue, exploring methods to obtain text embeddings from LLMs directly is a promising direction. In Table \ref{tab:embedders}, we list 40 LLM-based embedders and their detailed information. Specifically, we focus on their backbone selection, architectural improvement, training methods, and evaluation protocols.

\begin{landscape}
\tiny
    \begin{xltabular}{\textwidth}{lllllllllll}
    \caption{The overview of LLM-based embedders. The {\bf Paradigm} column is shortened as follows: supervised contrastive learning (SCL), weak supervised contrastive learning (WCL), unsupervised contrastive learning (UCL), training-free (TF), supervised next token prediction (SNTP), unsupervised next token prediction (UNTP), FLOPS regularization (FLOPS), masked next token prediction (MNTP), mask language modeling (MLM), Kullback-Leibler divergence (KL), metric learning (ML), auto-encoding (AE), iterative contrastive refinement (ICR) and reinforcement learning (RL).}\label{tab:embedders} \\
    \toprule
    \multirow{2}{*}{\bf Method} & {\bf Model} & \multicolumn{4}{c}{\bf Architecture} & \multicolumn{2}{c}{\bf Training} & \multicolumn{2}{c}{\bf Evaluation} \\
    \cmidrule(r){2-2} \cmidrule(r){3-6} \cmidrule(r){7-8} \cmidrule(r){9-10}
     & LLM (Encoder) & Pooling & Attention & Projector & PEFT & Input Format & Paradigm & Setting & Task  \\
    \midrule
    \endfirsthead

    \multicolumn{10}{l}{\footnotesize Continued on next page.}\\
    \toprule
    \multirow{2}{*}{\bf Method} & {\bf Model} & \multicolumn{4}{c}{\bf Architecture} & \multicolumn{2}{c}{\bf Training} & \multicolumn{2}{c}{\bf Evaluation} \\
    \cmidrule(r){2-2} \cmidrule(r){3-6} \cmidrule(r){7-8} \cmidrule(r){9-10}
     & LLM (Encoder) & Pooling & Attention & Projector & PEFT & Input Format & Paradigm & Setting & Task  \\
    \midrule
    \endhead

    \midrule
    \multicolumn{10}{l}{\footnotesize (to be continued on next page)}\\
    \endfoot

    \bottomrule
    \endlastfoot
    
    Sentence-T5 \cite{ni2022sentence} & T5 & Frist / Mean & Bi-Dir & $\surd$ & $\times$ & - & WCL→SCL & ZS / FT & STS / Clf.  \\ 
    PromptEOL \cite{jiang2023scaling} & \opt, \llama & Last & Causal & $\times$ & $\times$ & Prompt & TF / SCL & ZS / ICL & STS / Clf.  \\ 
    MetaEOL \cite{lei2024meta} & \llama, \mistral & Last & Causal & $\times$ & $\times$ & Prompt & TF & ZS / FT & STS / Clf.  \\ 
    PromptSTH/SUM \cite{zhang2024simple} & \opt, \llama, \mistral & Last & Causal & $\times$ & $\times$ & Prompt & TF & ZS & STS  \\ 
    Token Prepending \cite{fu2024token} & \llama, \qwen, Gemma & Last & Causal & $\times$ & $\times$ & Prompt & TF & ZS / FT & STS / Clf.  \\ 
    BeLLM \cite{li2024bellm} & \llama & Last & Bi-Dir & $\times$ & LoRA & Prompt & SCL & ZS / FT & STS / Clf.  \\ 
    AutoRegEmbed \cite{deng2025following} & \llama, \mistral & Special Seq Mean & Causal & $\times$ & $\times$ & Instruction & SNTP→SCL & ZS & STS  \\ 
    \midrule
    GTR \cite{ni2022large} & T5 & Mean & Bi-Dir & $\surd$ & $\times$ & - & WCL→SCL & ZS / FT & IR  \\ 
    SGPT-IR \cite{muennighoff2022sgpt} & \gptneo & Weighted Mean & Causal & $\times$ & Bi-DirtFit & - & SCL & ZS / FT & IR  \\ 
    Llama2Vec \cite{li2024llama2vec} & \llama & Special Last & Causal & $\times$ & LoRA & - & AE+UNTP→SCL & FT & IR  \\ 
    RepLLaMA \cite{ma2024fine} & \llama & Special Last & Causal & $\times$ & LoRA & - & SCL & ZS / FT & IR  \\ 
    BMRetriever \cite{xu2024bmretriever} & Pythia, Gemma, Bi-DiroMistral & Special Last & Causal & $\times$ & LoRA & Instruction & WCL→SCL & ZS / FT & IR  \\ 
    ChatRetriever \cite{mao2024chatretriever} & \qwen & Special Seq Last & Customized & $\times$ & LoRA & Instruction & SCL+MLM & ZS / FT & IR  \\ 
    PromptReps \cite{zhuang2024promptreps} & \mistral, Phi-3, \llama & Last & Causal & $\surd$ & $\times$ & Instruction+Prompt & TF / SCL & ZS & IR  \\ 
    LMORT \cite{sun2024llm} & GPT2, GPT-J & Multi-Layer & Causal & $\times$ & $\times$ & - & SCL & FT & IR  \\ 
    Mistral-SPLADE \cite{doshi2024mistral} & \mistral & Post & Causal & $\surd$ & QLoRA & Prompt & SCL+FLOPS & ZS & IR  \\ 
    NV-Retriever \cite{moreira2024nv} & \mistral & Mean & Bi-Dir & $\times$ & LoRA & Instruction & SCL→SCL & ZS & IR  \\ 
    Promptriever \cite{weller2025promptriever} & \llama & Special Last & Causal & $\times$ & LoRA & Instruction & SCL & ZS / FT & IR  \\ 
    RARe \cite{tejaswi2024rare} & \llama, \mistral & Mean / Last & Bi-Dir / Causal & $\times$ & LoRA & Instruction+Example & SCL & ICL & IR  \\ 
    DEBATER \cite{ji2025learning} & MiniCPM & Special Seq Last & Causal & $\times$ & LoRA & Instruction & SCL+KL & ZS & IR  \\
    O1 Embedder \cite{yan2025o1} & \mistral, \llama, \qwen & Gen. Seq Last & Causal & $\times$ & LoRA & Instruction & SCL+SNTP & ZS & IR  \\ 
    Search-R3 \cite{gui2025search} & \qwen & Gen. Seq Last & Causal & $\times$ & LoRA & Instruction & SNTP+KL+SCL+ML→RL & ZS & IR  \\ 
    ReasonEmbed \cite{chen2025reasonembed} & \llama, \qwen & Special Last & Causal & $\times$ & LoRA & Instruction & SCL & ZS & IR  \\ 
    \midrule
    Echo \cite{springer2024repetition} & \mistral & Partial Mean & Causal & $\times$ & LoRA & Instruction+Prompt & TF / SCL & ZS & Uni.  \\ 
    GenEoL \cite{thirukovalluru2024geneol} & \mistral & Mean & Causal & $\times$ & $\times$ & Prompt & TF & ZS & Uni.  \\ 
    MoEE \cite{li2024your} & Deepseek, \qwen, OLMoE & Multi-Layer & Causal & $\times$ & $\times$ & Prompt & TF & ZS & Uni.  \\ 
    ReBA \cite{duan2025retrieval} & GPT-2, \llama & Multi-Layer & Causal & $\times$ & $\times$ & - & TF & ZS & Uni.  \\ 
    LLM2Vec \cite{behnamghader2024llm2vec} & \llama, \mistral & Mean & Bi-Dir & $\times$ & LoRA & Instruction & MNTP→UCL/SCL & ZS & Uni.  \\ 
    Cpt \cite{neelakantan2022text} & Unknown & Special Last & Unknown & $\times$ & $\times$ & - & SCL & ZS & Uni.  \\ 
    UDEVER \cite{zhang2023language} & \bloom & Special Last & Causal & $\times$ & Bi-DirtFit & - & SCL & ZS & Uni.  \\ 
    InstructOR \cite{su2023one} & \gtr & Mean & Bi-Dir & $\surd$ & $\times$ & Instruction & SCL & ZS & Uni.  \\ 
    InBedder \cite{peng2024answer} & \opt, \llama & Last & Causal & $\times$ & $\times$ & Instruction+Prompt & SNTP & ZS & Uni.  \\ 
    GritLM \cite{muennighoff2024generative} & \mistral & Mean & Customized & $\surd$ / $\times$ & $\times$ & Instruction & SCL+SNTP & ZS / FS & Uni.  \\ 
    MLTP \cite{tang2024pooling} & \mistral & Multi-Layer & Bi-Dir / Causal & $\times$ & $\times$ & Instruction & SCL & ZS & Uni.  \\ 
    ULLME \cite{man2024ullme} & \llama, \mistral, \phii & Mean & Bi-Dir / Causal & $\times$ & LoRA & - & SCL+RL+KL & ZS & Uni.  \\ 
    L$^3$Prune \cite{fischer2024large} & \llama, \mistral, \qwen, \phii & Weighted Mean & Causal & $\times$ & LoRA & Instruction & SCL & ZS & Uni.  \\ 
    LENS \cite{lei2025enhancing} & \mistral & Post & Bi-Dir & $\times$ & LoRA & Instruction & SCL & ZS & Uni.  \\ 
    DIFFEMBED \cite{zhang2025diffusion} & Dream & Mean & Bi-Dir & $\times$ & LoRA & Instruction & SCL & ZS & Uni.  \\ 
    MGH \cite{pan2025negative} & \mistral & Weighted Mean & Bi-Dir & $\times$ & LoRA & Instruction & SCL & ZS & Uni.  \\ 
    GRACE \cite{sun2025grace} & \llama, \qwen & Hyb. Seq Mean & Causal & $\times$ & $\times$ & Instruction & RL & ZS & Uni.  \\
    Lychee \cite{zhang2025phased} & \qwen & Special Last & Causal & $\times$ & LoRA & Instruction & SCL→SCL→MG→SCL & ZS & Uni.  \\ 
    GIRCSE \cite{tsai2025let} & \mistral, \qwen & Gen. Seq Mean & Causal & $\times$ & LoRA & Instruction & SCL+ICR & ZS & Uni.  \\ 
    Anchor \cite{su2025training} & \mistral, \llama, \qwen & Special Last & Causal & $\times$ & LoRA & Instruction & SNTP→SCL & ZS & Uni.  \\ 
    Text2Token \cite{an2025text2token} & \llama, \mistral & Mean / Last & Bi-Dir / Causal & $\times$ & LoRA & - / Prompt & UNTP & ZS & Uni.  \\ 
    \midrule
    \multicolumn{10}{c}{\it Series works by commercial companies}  \\
    \midrule
    CoDiEmb (Tecent) \cite{zhang2025codiemb} & MiniCPM & Mean & Causal & $\times$ & $\times$ & Instruction & SCL+Pearson+KL+RL→MG & ZS & Uni.  \\ 
    Conan-Embedding-v2 (Tecent) \cite{li2025conan} & 1.4B LLM & Mean & Causal→Bi-Dir & $\times$ & $\times$ & Instruction & UNTP→SNTP→WCL→SCL & ZS & Uni. \\
    F2LLM (Ant) \cite{zhang2025f2llm} & \qwen & Special Last & Causal & $\times$ & $\times$ & Instruction & SCL & ZS & Uni.  \\ 
    Linq-Embed-Mistral (Linq AI) \cite{choi2024linq} & \mistral & Special Last & Causal & $\times$ & LoRA & Instruction & SCL & ZS & Uni.  \\ 
    QZhou-Embedding (KingSoft) \cite{yu2025qzhou} & \qwen & Mean & Bi-Dir & $\times$ & $\times$ & Instruction & SCL→SCL+ML & ZS & Uni.  \\
    \midrule
    {\bf Flag Embedding (BAAI)} &  &  &  &  &  &  &  &  &   \\ 
    BGE-ICL \cite{li2024making} & \mistral, \gemma & Special Last & Bi-Dir / Causal & $\times$ & LoRA & Instruction+Example & SCL & ZS / ICL & Uni.  \\ 
    BGE-Multilingual-Gemma2 & \gemma & Special Last & Causal & $\times$ & $\times$ & Instruction & Unknown & ZS & Uni.  \\
    \midrule
    {\bf E5 Embedding (Microsoft)} &  &  &  &  &  &  &  &  &   \\ 
    E5-Mistral \cite{wang2024improving} & \mistral & Special Last & Causal & $\times$ & LoRA & Instruction & SCL & ZS & Uni.  \\ 
    SPEED \cite{chen2024little} & \mistral & Special Last & Causal & $\times$ & LoRA & Instruction & SCL & ZS &   \\
    \midrule
    {\bf Gemini Embedding (Google)} &  &  &  &  &  &  &  &  &   \\ 
    Gecko \cite{lee2024gecko} & Unknown & Mean & Unknown & $\times$ & $\times$ & Instruction & WCL→SCL & ZS & Uni.  \\ 
    Gemini Embedding \cite{lee2025gemini} & Gemini & Mean & Bi-Dir & $\surd$ & $\times$ & Instruction & WCL→SCL→MG & ZS & Uni.  \\
    \midrule
    {\bf (Nvidia)} &  &  &  &  &  &  &  &  &   \\ 
    NV-Retriever \cite{moreira2024nv} & \mistral & Mean & Bi-Dir & $\times$ & LoRA & Instruction & SCL→SCL & ZS & IR  \\ 
    NV-Embed-v1 \cite{lee2024nv} & \mistral & Post & Bi-Dir & $\times$ & LoRA & Instruction & WCL→SCL & ZS & Uni.  \\ 
    NV-Embed-v2 & \mistral & Post & Bi-Dir & $\times$ & Unknown & Instruction & WCL→SCL & ZS & Uni.  \\ 
    \midrule
    {\bf Qwen Embedding (Alibaba)} &  &  &  &  &  &  &  &  &   \\ 
    GTE-Qwen2 \cite{li2023towards} & \qwen & Special Last & Bi-Dir & $\times$ & $\times$ & Instruction & WCL→SCL & ZS & Uni.  \\ 
    Qwen3 Embedding \cite{zhang2025qwen3} & \qwen & Special Last & Causal & $\times$ & $\times$ & Instruction & WCL→SCL→MG & ZS & Uni.  \\ 
    \midrule
    {\bf SFR-Embedding (Salesforce)} &  &  &  &  &  &  &  &  &   \\ 
    SFR-Mistral \cite{SFRAIResearch2024} & \mistral & Special Last & Causal & $\times$ & LoRA & Instruction & SCL & ZS & Uni.  \\ 
    SFR-Embedding-2 & \mistral & Special Last & Causal & $\times$ & $\times$ & Instruction & Unknown & ZS & Uni.  \\ 
    \end{xltabular}
\end{landscape}

\subsection{Backbone Selection}
\paragraph{Model} The LLM-based embedders appearing in Table \ref{tab:embedders} use numerous open-source LLMs as backbones. Among the encoder-decoder backbones, T5 (GTR) has a dominant position, which is not surprising, as T5 and its variants have been leading innovations in encoder-decoder PLMs. Among decoder-only backbones, the most popular is Mistral (29 / 69), followed by LLaMA (21 / 69) and Qwen (15 / 69). The popularity of the Mistral is strongly connected to its excellent performance: a recent study \cite{behnamghader2024llm2vec} find that Mistral's embedding performance will not drop significantly compared to the other LLMs when converting causal attention to bi-directional attention. This characteristic may derive from its pre-training or instruction fine-tuning phase, but no information has been revealed. Specifically, DIFFEMBED~\cite{zhang2025diffusion} employs the diffusion language model \texttt{Dream 7B}~\cite{ye2025dream} as its backbone, demonstrating exceptional potential.

\paragraph{Parameter Size} The vast majority of LLM-based embedders have a backbone size set to 7B, which, directly speaking, may result from a trade-off between efficiency and performance. However, the only several attempts to use LLM with larger LLMs lead to even worse performance \cite{muennighoff2024generative,jiang2023scaling}. Therefore, the scaling law of LLM-based embedders need to be explored further.

\subsection{Architecture Improvement}
Denoting $d$ as the dimension of hidden states, $L$ as the number of transformer layers, and $\mathcal{V}$ as the vocabulary size, without loss of generality, we divide the LLM $F$ into multiple Transformer layers and a decoding layer, which is denoted as
\begin{align*}
    F = g \circ [ f^{(L)} \circ \cdots \circ f^{(1)} ] \circ f^{(0)} = g \circ f
\end{align*}
where $f^{(0)}: \mathbb{R}^{|\mathcal{V}|} \to \mathbb{R}^d$ is the input layer, $f^{l}: \mathbb{R}^d \to \mathbb{R}^d$ is the $l$-th transformer layer and $g: \mathbb{R}^d \to \mathbb{R}^{|\mathcal{V}|}$ is the decoding layer. Although some studies~\cite{oh2022don,li2024bellm} have found that the optimal embeddings do not (solely) originate from the final layer, the vast majority of current approaches still derive embeddings from the pooling of hidden states from the final layer. We use $f$ to represent $[ f^{(L)} \circ \cdots \circ f^{(1)} ] \circ f^{(0)}$ for convenience.

\subsubsection{Traditional Pooling Strategy}\label{sec:embedder_pooling}
Similar to traditional language models, the LLM includes a decoding layer that completes the mapping from the hidden state space to the token vocabulary space. To the best of our knowledge, almost all LLMs adhere to the original design of GPT \cite{radford2018improving}, utilizing a simple, unbiased linear layer as the decoding layer. When LLMs are used as textual embedders, embeddings are obtained by a specific pooling strategy at hidden states, while mapping to the token space is no longer necessary. Therefore, most methods directly discard the decoding layer and opt for a pooling strategy on the hidden state of the last transformer layer's output to obtain a single embedding for each text. To derive a universal formula, we introduce some optional components: (1) Instruction $\gI$; (2) In-Context Examples $\gE$; (3) Prompt Template \texttt{prompt($\cdot$)}; (4) Special Token Sequences $\gS$. Then we can express the output of the LLM-based embedder as
\begin{align}
    [\vh^{(L)}_0, \vh^{(L)}_{\gI_1}, \cdots \vh^{(L)}_{\gI_m},\vh^{(L)}_{\gE_1}, \cdots \vh^{(L)}_{\gE_t}, \vh^{(L)}_{x_1}, \cdots \vh^{(L)}_{x_n}, \vh^{(L)}_{\gS_1}, \cdots, \vh^{(L)}_{\gS_k}] = \mathcal{H} = f(\gI \oplus \gE \oplus {\rm prompt}(x) \oplus \gS),
\end{align}
where $\vh^{(L)}$ is the $L$-layer hidden state of $f$; $m$, $t$, $n$, and $k$ denote the number of tokens in $\gI$, $\gE$, ${\rm prompt}(x)$, and $\gS$, respectively. Specially, $\vh^{(L)}_0$ represents the last hidden state corresponding to the starting token (e.g., \texttt{[<pad>]} for T5 or \texttt{[<s>]} for LLaMA) added by the models by default. Although some works \citep{moreira2024nv,behnamghader2024llm2vec,muennighoff2024generative,su2023one} splice instructions or in-context examples in front of the input text, obviously, its hidden states are not considered during the pooling stage. Therefore, we can expressed the pooling strategy as $\mathcal{P}(\cdot)$:
\begin{align}\label{eqn:pooling}
    \vh = \operatorname{\mathcal{P}}([\vh^{(L)}_0, \vh^{(L)}_{x_1}, \cdots \vh^{(L)}_{x_n}, \vh^{(L)}_{s_1}, \cdots, \vh^{(L)}_{s_k}]),
\end{align}
where $n$ is the token number of the input $x$, depending on the tokenizer of the LLMs. In this section, we have categorized the pooling strategy into six distinct types: (1) First Pooling; (2) Mean Pooling; (3) Last Pooling; (4) Post-Interaction Pooling; (5) Multi-Layer Pooling; (6) Generative Pooling. Each method will be introduced below separately.

\paragraph{First Pooling} First pooling is a common way of obtaining an embedding for PLMs with bi-directional attention, which can be express as
\begin{align}
    \vh = \operatorname{\mathcal{P}}([\vh^{(L)}_0,\vh^{(L)}_{x_1}, \cdots \vh^{(L)}_{x_n}, \vh^{(L)}_{s_1}, \cdots, \vh^{(L)}_{s_k}]) = \vh^{(L)}_0
\end{align}
For example, BERT splices the special token \texttt{[CLS]} before the input text, and the embedding output from the \texttt{[CLS]}'s position will be used as the embedding for the whole text \cite{reimers2019sentence}. Although T5-like models do not have the special token like \texttt{[CLS]} in BERT, Sentence-T5 \cite{ni2022sentence} use the first embedding output by T5's encoder and the \texttt{[START]} embedding output by T5's decoder as the embedding of the whole text; however, these pooling strategies obtain slightly worse performance than mean pooling. For the decoder-only LLMs with causal attention, the first pooling does not work, as the embedding of the first position can't be included in the semantics of the subsequent content.

\paragraph{Mean Pooling} 
Traditional Mean pooling averages the embedding of each position and shows better performance in BERT than first pooling \cite{reimers2019sentence,ni2022sentence}.

\begin{align}\label{eqn:mean_pooling}
    \vh = \operatorname{\mathcal{P}}([\vh^{(L)}_{x_1}, \cdots \vh^{(L)}_{x_n}, \vh^{(L)}_{s_1}, \cdots, \vh^{(L)}_{s_k}]) = \sum_{i=1}^{n} \alpha_i \vh_{x_i} + \sum_{j=1}^{k} \beta_j \vh_{s_j}
\end{align}

\begin{itemize}
    \item {\bf Naive Mean Pooling (Mean)} The naive mean pooling generally does not include special tokens and $\forall i,j$ satisfies $\alpha_i = 1/n,\ \beta_j = 0$ in Eqn \ref{eqn:pooling}. Up to now,, mean pooling remains regarded as one of the most straightforward methods, widely employed across various LLM backbones.
    
    \item {\bf Weighted Mean Pooling (Weighted Mean)} The weighted mean pooling fits $\alpha_i = i / \sum_{i=1}^ni,\ \beta_j = 0$ for $\forall i,j$ in Eqn \ref{eqn:mean_pooling}. For decoder-based LLMs, assigning the same weight to each position because the latter position will be able to see more semantic information due to causal attention \cite{muennighoff2022sgpt}. SGPT \cite{muennighoff2022sgpt} suggests the use of weighted mean pooling with the intuition that the latter position should be given a larger weight. L$^3$Prune \cite{fischer2024large} follows SGPT's pooling strategy to fine-tune the pruned LLM-based embedders.
    
    \item {\bf Partial Mean Pooling (Partial Mean)} The partial mean pooling assign $\alpha_i = 1/l$ for $\forall i \in [k, k+l]$ while the other elements is 0 in Eqn \ref{eqn:mean_pooling}. $k$ is an example-specific number and $l$ is the token number of the input text $x$. This strategy is viewed as an effective method to mitigate the effects of causal attention, which requires prompt-based assistance. Echo \cite{springer2024repetition} propose a prompt: \texttt{Rewrite the sentence:[x], rewritten sentence:[x]}, where \texttt{[x]} is the placeholder. In practice, both placeholders are filled with the same text, and the mean pooling strategy is used to obtain the text embedding, but it is pooled only within the range of the second occurrence of the text. In this way, the entire text is already present in the first placeholder, so each token populated in the second placeholder can access the entire text information through causal attention. Mistral-SPLADE \cite{doshi2024mistral} follows Echo's approach and uses the obtained embedding to generate sparse representations.

    \item {\bf Special-Sequence Mean Pooling (Special Seq Mean)} The special-sequence mean pooling can be denoted as $\alpha_i =0, \beta_j = 1/k$ for $\forall i, j$. in Eqn.\ref{eqn:mean_pooling}. AutoRegEmbed \cite{deng2025following} is the only method employing this pooling strategy, which draws inspiration from long text compression techniques. The authors introduce the document prediction task, guiding the embedder condenses the semantic information of the entire text into multiple special tokens. Then the mean pooling on all special tokens aggregates the information from these tokens to obtain the text embedding.
\end{itemize}

\paragraph{Last Pooling} Last pooling is a new pooling strategy emerging in the LLM era. Due to the unidirectionality of causal attention, only the last position shows information about the whole text. However, decoder-only LLMs are pre-trained by the next token prediction, and the embedding at the last position will align with the potential next token embedding \cite{jiang2023scaling}. Without additional interventions, there is no guarantee that the embedding contains the semantics of the entire text. Therefore, last pooling ultimately requires the use of prompts and/or special tokens to achieve semantic aggregation, which can be expressed as follows in the technique:
\begin{align}
    \vh = \operatorname{\mathcal{P}}([\vh^{(L)}_{x_1}, \cdots \vh^{(L)}_{x_n}, \vh^{(L)}_{s_1}, \cdots, \vh^{(L)}_{s_k}]) = \left\{
    \begin{aligned}
        & \vh^{(L)}_{x_n}, \quad \# \text{prompt-based last pooling} \\
        & \vh^{(L)}_{s_k}, \quad \# \text{special-token / special-sequence last pooling} \\
    \end{aligned}
    \right.
\end{align}
\begin{itemize}
    \item {\bf Prompt-Based Last Pooling (Last)}. A special prompt is introduced to induce the model to summarize the semantics of the whole text in the last position \cite{jiang2023scaling,zhang2024simple,li2024bellm,zhuang2024promptreps,thirukovalluru2024geneol,li2024your}. For example, PromptEOL \cite{jiang2023scaling} propose a prompt template \texttt{The sentence [X] means in a word:``}, where \texttt{[X]} is a placeholder. In practice, \texttt{[X]} is filled with the input text, and the last pooling strategy is used to obtain the text embedding. Although there are many variants, the core idea in the line of work is consistent, i.e., convert text embedding to language modelling and guide the model to summarize the whole text semantics in the last position. These methods are marked as ``last'' in the ``pooling'' column of Table \ref{tab:embedders}.
   
    \item {\bf Special-Token Last Pooling (Special Last)}. Some works~\cite{neelakantan2022text,zhang2023language,li2024llama2vec,zhuang2024promptreps,mao2024chatretriever,xu2024bmretriever,ma2024fine} introduce a special token, such as \texttt{<EOS>}, to obtain embedding in the last position, while the LLM is fine-tuned incrementally, learning to converge semantics at the position of the special token.
    
    \item {\bf Special-Sequence Last Pooling (Special Seq Last)}. ChatRetriever~\cite{mao2024chatretriever}, DEBATER~\cite{ji2025learning} and AutoRegEmbed \cite{deng2025following} add a special token sequence, such as $[\texttt{EMB}_1], \cdots,[\texttt{EMB}_t]$, at the end of the input text. The authors argue that these $t$ consecutive special tokens serve as a chain of thought that extends and guides the learning space for more effective embeddings.
\end{itemize}

\paragraph{Post-Interaction Pooling} Some works add complex modules with attention mechanisms before the conventional pooling strategy, where $\mathcal{P}(\cdot)$ is a learnable network. NV-Embed \cite{lee2024nv} introduces an additional attention-based network before mean pooling to post-interact with the hidden states $\mathcal{H}$. The attention-based network comprises a cross-attention block and an MLP, where the key and value matrices used in the cross-attention are learnable and low-rank, and $\mathcal{H}$ is considered the query. Mistral-SPLADE \cite{doshi2024mistral} uses the same prompt as Echo \cite{springer2024repetition} to obtain text embeddings and follows the SPLADE recipe \cite{formal2021splade} to fine-tune the Mistral-based embedding. LENS \cite{lei2025enhancing} further clusters the LLM's tokens according to their embedding to narrow the sparse representation latitude and reduce information redundancy, achieving better sparsification results.

\subsubsection{Special Pooling Strategy}

Some novel approaches explore methods that do not (solely) rely on the hidden states of the final layer. Therefore, they cannot be expressed using Eqn \ref{eqn:pooling}. We categorize these methods into two types: (1) Multi-layer Pooling, which attempts to enhance embeddings by utilizing information contained in outputs from other layers of the model; (2) Generative Pooling, which seeks to incorporate the content generated by LLMs into the pooling process. Below, we will provide a detailed introduction to each method:

\paragraph{Multi-Layer Pooling} Some works mix the information from the different layers' hidden states in the LLM. For example, MLTP \cite{tang2024pooling} proposes a trainable multi-layer pooling method that aggregates text embeddings from multiple layers using last/mean pooling. These embeddings form a matrix $\mathcal{H}^{L}$, which is processed by a cross-attention Transformer. Cross-attention is computed by summing $\mathcal{H}^{L}$ with a learnable weight matrix, then deriving key and value representations. A fixed learnable query is used across inputs, and the feed-forward block output serves as the final embedding. \cite{sun2024llm} collects the hidden states from the layer with the optimal \textit{alignment} and \textit{uniformity} metric \cite{wang2020understanding}, which denotes $\mathcal{H}^a$ and $\mathcal{H}^u$ separately. Then, a multi-layer network called LMORT is introduced to fuse the information in both $\mathcal{H}^a$ and $\mathcal{H}^u$, where each layer consists of a bi-directional self-attention block, a bi-directional cross-attention block, a feed-forward block, and an inter-block residual mechanism. The authors demonstrate that the performance of retrieval tasks can be enhanced by fine-tuning the LMORT's parameters alone. MoEE \cite{li2024your} utilizes the routing weights from Mixture-of-Experts (MOE) LLMs as embeddings, demonstrating a promising complementarity with hidden states from the last layers. ReBA \cite{duan2025retrieval} aggregates the weights of all attention heads in different layers to recompute the final token embedding while utilizing a similar ``repetition'' idea proposed by Echo \cite{duan2025retrieval}.

\paragraph{Generative Pooling} Some works enhance embedding quality through multi-step reasoning or augment the original input using generated components; therefore, they incorporate the generated part by LLMs into pooling strategies. 

\begin{itemize}
    \item {\bf Generated Sequence Mean Pooling (Gen. Seq Mean)} GIRCSE~\cite{tsai2025let} sets an iterative step count $k$, and uses the average of the final hidden states at the generated $k$ positions as the text embedding.
    \item {\bf Generated Sequence Last Pooling (Gen. Seq Last)} O1 Embedder~\cite{yan2025o1} and Search-R3~\cite{gui2025search} employ supervised fine-tuning and reinforcement learning approaches respectively to enable the LLM to perform reasoning first, ultimately generating a special token to obtain the embedding.
    \item {\bf Hybrid Sequence Mean (Hyb. Seq Mean)} GRACE~\cite{tsai2025let} averages the last hidden states of both input and output content, excluding the instruction text, which is regarded as the text embedding.
\end{itemize}

\subsubsection{Attention Mechanism}\label{sec:attention}
The discussion of attention in the current approach focuses on a decoder-only LLM-based embedder. Causal Attention \cite{ashish2017attention} is a common operation for decoder-based LLMs, which ensures that the language modeling can only refer to the prefix to predict the next token. However, extensive empirical studies \cite{li2023label,li2024bellm,springer2024repetition} have shown that causal attention will degrade downstream task performance.  Current approaches either (1) retain the causal attention and use other tricks (e.g., special pooling methods) or (2) convert it to bi-directional attention and make the model adapt to it.

\paragraph{Causal Attention}
If the model maintains causal attention, it is often necessary to resort to pooling tricks to mitigate the resulting adverse effects. Representatives of these methods include SGPT \cite{muennighoff2022sgpt}, PromptEOL \cite{jiang2023scaling}, and Echo \cite{springer2024repetition}; please refer to Section \ref{sec:embedder_pooling} for details.

\paragraph{Bi-Directional Attention} Considering that the causal attention is implemented by a mask matrix implementation in Transformer \footnote{Transformers (\url{https://github.com/huggingface/transformers}) is the most popular open-source toolkit to instantiate LLMs.}, it is convenient and quick to remove the mask matrix during incremental fine-tuning. BeLLM \cite{li2024bellm} pioneered an attempt to change the causal attention of the last few layers in the decoder-only LLM to bi-directional attention. This trial is motivated by the fact that, as the number of layers increases, the performance of downstream tasks does not improve monotonically, but there is a turning point. When fully converting the causal attention in decoder-only LLM to bi-directional, sufficient data or additional training tasks are required to adapt the LLM to the new attentional mechanism. The experimental results in \cite{li2024bellm} show that when training on small-scale data, changing the attention of only the last layer from causal to bi-directional can effectively improve performance on the SemEval Benchmark; however, directly changing all the attentional layers to bi-directional results in a significant decrease in performance.
Furthermore, NV-Embed \cite{lee2024nv} demonstrates that LLMs can adaptively update their parameters and complete the conversion from causal attention to bi-attention without requiring additional operations, provided sufficient data (hundreds of datasets) is available for incremental fine-tuning. To adapt the LLM to use bi-directional attention without massive training data, LLM2Vec \cite{behnamghader2024llm2vec} proposes a self-supervised task, masked next token prediction (MNTP). MNTP combines the ideas of mask language modeling and next token prediction, where the token is first randomly masked from the text, and the hidden state from the previous position of the mask token is used to predict the masked token. In addition, Conan-Embedding-v2~\cite{li2025conan} proposes a novel approach: smoothly transitioning from causal attention to bidirectional attention during training.

\paragraph{Dynamic Transformation} GirtLM \cite{muennighoff2024generative} is capable of both generation and embedding tasks with a multi-task learning paradigm (Please see Section \ref{sec:multi-task} for details). Therefore, GirtLM can shift to bi-directional attention for high-quality embedding while maintaining causal attention for generation, which is consistent with the training setting.

\subsubsection{Additional Projector}
Adding projectors is the usual strategy in text embedding. Unlike pooling strategies that operate on multiple hidden states, projectors usually operate on a single embedding for other purposes. We divide the existing work into three parts based on the projectors' purposes.

\paragraph{Projector for Low-Dimensional Embedding} The hidden layer dimension (width) usually increases with the number of total parameters, according to the scaling laws \cite{kaplan2020scaling}. The hidden states of T5-11B have 1024 dimensions, whereas the mainstream 7B decoder-only LLMs have 4096 dimensions, both of which are higher than the 768 dimensions of a number of BERT-based models. In practice, high-dimensional embedding dramatically increases the resource overhead for storage and inference, and a simple approach is to add a projector $g: \mathbb{R}^d \to \mathbb{R}^m, m < d$  after pooling to obtain the low-dimensional embedding. With a fixed output dimension linear layer, Sentence-T5 \cite{ni2022sentence} and GTR \cite{ni2022large} empirically demonstrate that the performance on STS, classification, and retrieval tasks can be improved by scaling the embedder's parameter only without scaling the dimension. GirtLM \cite{muennighoff2024generative} also uses a linear layer to map Mistral's 4096-dimensional embedding to 1024 and finds that the performance of MTEB would be slightly degraded.

\paragraph{Projector for Sparse Representation} Similar to the purpose of dimensionality reduction, converting text embedding to sparse representations can improve the inference efficiency of the model \cite{formal2021splade} and enhance its performance on partial downstream tasks such as long document retrieval \cite{chen2024bge}. For BERT-based embedders, text embeddings can be mapped to the vocabulary-length logits through the projector for mask language modeling; then, the vocabulary-size logits can be regarded as the text representation. To sparsify the representation, common methods include gate mechanisms \cite{bai2020sparterm}, top-k masking \cite{yang2021sparsifying}, and regularization terms \cite{paria2020minimizing}. Similarly, text embeddings from LLM-based embedders can go through the decoder layer and use a similar idea for sparsification. Specifically, PromptReps \cite{zhuang2024promptreps} uses the prompt similar to PromptEOL to obtain text embeddings and introduce ReLU function, log-saturation \cite{formal2021splade} and top-k masking to obtain the sparse representation for documents. In addition, an empirical study \cite{nie2024text} finds that the embedding output from LLM-based embedders, which is fine-tuned by contrastive learning, still produces high-quality sparse representations via the decoding layer and simple top-$k$ masking.

\paragraph{Projector for Customized Adaptation} 
Search-Adaptor~\cite{yoon2024search} proposes the use of additional projectors for customizing commercial text embedding interfaces, improving performance on private domain data.  Furthermore, Matryoshka-Adaptor~\cite{yoon2024matryoshka} and SMEC~\cite{zhang2025smec} attempt to obtain embeddings across different dimensions while customizing the adaptation, and maintain the quality of low-dimensional embeddings by employing distinct optimization objectives.

\subsubsection{Parameter-Efficient Fune-Tuning Module}
Recall that the current parameter scales used for treating LLM-based Embedder are around 7B, e.g., both Mistral-7B and LLaMA3-8B are the most commonly used backbones. Under the popular half-precision (TF16 or BF16) setting, one 7B LLM can occupy $\sim$60GB RAM for full-parameter fine-tuning or $\sim$16G RAM for parameter-efficient fine-tuning using LoRA, which both can be accomplished on two NVIDIA V100 / A100 GPUs. Thus, whether or not parameter-efficient fine-tuning techniques are used depends primarily on the amount of data used, not on resource constraints. Some works introduce BitFit \cite{zaken2022bitfit}, or LoRA \cite{hu2021lora} as the PEFT modules and fine-tune using a single dataset, such as Wiki1M \cite{behnamghader2024llm2vec}, SNLI \cite{muennighoff2022sgpt}, or MSMARCO \cite{li2024llama2vec}, in the belief that a small-size data can stimulate the semantic generalization capabilities that LLMs themselves possess. The other works use full parameter tuning and (multiple-stage) fine-tuning based on hundreds of datasets, which was sufficient to allow large changes to the model parameters and avoid overfitting.

\subsection{Optimization}

With the advent of LLMs, the landscape of text embeddings has significantly evolved. Optimizing these embeddings is essential to enhance their quality, applicability, and efficiency across diverse applications.

\subsubsection{Training Free (TF)}
Some works leverage the inherent capabilities of large language models without additional fine-tuning. By carefully crafting input prompts, practitioners can guide the model to produce embeddings that are desirable and tailored to specific tasks. In the era of PLM, PromptBERT~\cite{jiang2022promptbert} first proposed a prompt-based method to enhance the text representation of the BERT model in training-free setting. Specifically, this method uses a fixed template (e.g., "[X] is [mask]", where [X] is a placeholder) and leverages the hidden states at the mask position as the sentence representation.  Recently, two lines of methods for LLMs have been developed:

\begin{itemize}
    \item {\bf Summary-Style Prompt}. Several approaches leverage prompt engineering to condense input text into a single word, aiming to capture core semantics. For instance, PromptEOL \cite{jiang2023scaling} employs the template \texttt{``The sentence [X] means in a word:''} to elicit a one-word summary, where \texttt{[X]} is replaced with the input text. The final hidden state is then used as the sentence embedding. Furthermore, PromptEOL enhances performance by utilizing in-context learning, generating one-word summaries with GPT-4 for a sentence and prepending them to the input as contextual examples. Building on this, PromptReps \cite{zhuang2024promptreps} expands the utility of prompt engineering by not only crafting dense text embeddings but also enabling sparse bag-of-words representations derived from language models. To further enhance LLM embedding expressiveness, PromptSTH/SUM \cite{zhang2024simple} proposes a Pretended Chain of Thought and Knowledge Enhancement. Pretended Chain of Thought simulates step-by-step reasoning by prepending prompts with phrases like "After thinking step by step," encouraging detailed sentence representation. Knowledge Enhancement integrates human summarization expertise by incorporating relevant knowledge into prompts, guiding the model to focus on essential information. MetaEOL \cite{lei2024meta} guides LLMs to produce embeddings through a series of carefully designed prompts that address multiple representational aspects, then uses an average of these meta-task-derived embeddings as the final representation. Similarly, GenEOL \cite{thirukovalluru2024geneol} introduces a novel approach by generating diverse sentence variations of the target text using an LLM. Subsequently, it computes the final sentence embedding by averaging the embeddings of these generated variants. Different from these, MoEE \cite{li2024your} leverages the inherent structure of Mixture-of-Experts (MoE) LLMs, combining routing weights and hidden states to produce robust embeddings without requiring explicit fine-tuning.
    \item {\bf Repetition-Style Prompt}. The inherent causal attention mechanism in decoder-only LLMs can restrict information flow for embedding models, prompting research into mitigation strategies via prompt design. Echo et al. \cite{springer2024repetition} introduce a self-repetition prompt, \texttt{Rewrite the sentence:[x], rewritten sentence:[x]}, where \texttt{[x]} represents the input sequence. This method leverages the repetition of the input to enhance contextual understanding, extracting text embeddings through mean pooling applied solely to the second occurrence of \texttt{[x]}. ReBA \cite{duan2025retrieval} extends this concept by developing an algorithm that directly modifies the model's attention matrix, using the repeated input to refine hidden state representations at each position, leading to enhanced performance compared to basic repetition. Token Prepending \cite{fu2024token} proposes a plug-and-play, training-free approach. This technique involves prepending the decoded sentence embedding from each layer to the input of the subsequent layer. By doing so, it allows earlier tokens to access the complete sentence embedding, effectively circumventing the limitations imposed by causal attention.
\end{itemize}

\subsubsection{Unsupervised Contrastive Learning (UCL)}
Before the era of LLMs, research demonstrated that utilizing data augmentation for unsupervised contrastive learning effectively enhances the quality of embeddings \cite{gao2021simcse}. This unsupervised training paradigm eliminates the dependence on high-quality supervised data, improves training efficiency, and enhances scalability, thereby attracting significant attention from researchers. The emergence of LLMs, along with prompt engineering methods, can be treated as an even more efficient unsupervised learning paradigm. Nonetheless, in the era of LLMs, studies tend to employ Unsupervised Contrastive Learning Methods in embedding learning primarily as a step within multi-stage learning processes to bolster specific aspects of the final model, rather than relying solely on UCL to train the entire model as seen in previous works like SimCSE \cite{gao2021simcse}. For instance, LLM2Vec \cite{behnamghader2024llm2vec} explores masked next-token prediction and unsupervised contrastive learning methods~\cite{gao2021simcse} while enabling bi-directional attention to enhance embedding performance. 

\subsubsection{Supervised Contrastive Learning (SCL)}
Supervised contrastive learning leverages labeled data to guide the embedding optimization process. By combining the inherent world knowledge of LLMs with high-quality supervised data, the Supervised Contrastive Learning Method is currently the mainstream paradigm for constructing embedding models based on LLMs. The  improvements primarily focus on the following areas:

\paragraph{Prompt Tuning (Prompt)} Prompt Tuning \cite{schick2021exploiting} aligns the fine-tuning form of downstream tasks with that of the pre-training task, enabling the mode learned during pre-training to be fully utilized. Some prompt-based training-free methods, such as PromptEOL \cite{jiang2023scaling}, PromptReps \cite{zhuang2024promptreps}, and Echo \citep{springer2024repetition}, further, BeLLM \cite{li2024bellm} explores the impact of backward dependencies in LLMs and then proposes a backward dependency enhanced large language model (BeLLM) that transforms certain attention layers from unidirectional to bidirectional. This modification allows the model to learn more nuanced sentence embeddings, improving performance across various STS tasks and demonstrating the value of incorporating backward dependencies.

\paragraph{Instruction Tuning (Instruction)}

General-purpose embedding models confront inherent challenges stemming from the diverse and often conflicting objectives across various downstream tasks. For instance, a sentence pair deemed semantically similar may prove irrelevant in a retrieval context if it fails to address a specific query. Early approaches attempted to mitigate this issue by differentiating symmetric and asymmetric retrieval through the use of prefixes, such as 'query:' and 'passage:'. However, with the expanding landscape of applications, including classification, clustering, retrieval, and semantic similarity assessment, instruction-based prompting has emerged as a promising strategy. This approach aims to enhance task-specific performance by explicitly incorporating task descriptions within the input. Instructor~\cite{su2023one} firstly pioneered instruction tuning for embedding models, building upon GTR~\cite{ni2022large} and curating a dataset encompassing various tasks. Building upon this foundation, Data-CUBE~\cite{min2024data} advanced the field by implementing curriculum learning within multitask instruction tuning. TART~\cite{asai2023task} specifically targeted retrieval tasks, constructing the BERRI dataset, which covers 40 distinct retrieval scenarios. The robust instruction-following capabilities of large language models (LLMs) have catalyzed the development of LLM-based embedding models with instruction tuning. InBedder~\cite{peng2024answer} proposes a novel methodology for training instruction-following sentence embedders by exploiting the generative capacity of LLMs. Rather than directly manipulating embedding vectors, InBedder fine-tunes LLMs on abstractive question answering, extracting sentence embeddings from the averaged hidden states of generated answers during inference. E5-Mistral~\cite{wang2024improving} utilizes a large, LLM-generated, instruction-rich training dataset to tune LLM embeddings. Notably, this model initially applied instructions only 0to the query side, a practice subsequently adopted by numerous subsequent works~\cite{xu2024bmretriever,mao2024chatretriever,moreira2024nv,tejaswi2024rare,ji2025learning,behnamghader2024llm2vec,muennighoff2024generative,li2023towards,lee2024nv,tang2024pooling,fischer2024large,li2024making,choi2024linq}.

\paragraph{In-Context Tuning (Example)} The practice of in-context tuning~\cite{min2022metaicl}, which augments training inputs with contextual examples, has proven effective in enhancing a model's understanding of contextual dependencies. Following this trend, RARe~\cite{tejaswi2024rare} and BGE-ICL~\cite{li2024making} have explored the adaptation of in-context tuning for LLM-based embedding models. These approaches involve fine-tuning pre-trained LLMs with in-context examples that share semantic similarity with the target input, resulting in improved performance on various downstream tasks.

\subsubsection{Next Token Prediction}
Next Token Prediction is a primary paradigm in the pre-training and supervised fine-tuning of LLMs. Obviously, the supervisory signal for next token prediction resides in the vocabulary space, inherently resulting in a dimensionality gap with the embedding space. This makes it challenging to design supervision signals in the vocabulary space and correctly influence the quality of text embeddings. Although theoretical research in this area remains scarce, interestingly, several successful empirical cases have demonstrated the potential of this generative task in text embedding. Note that the works introduced in this section replace contrastive learning with next token prediction as the primary task. For the work that regards next token prediction as a pretext task before contrastive learning, please refer to Section \ref{sec:multi-stage}.

\paragraph{Unsupervised Next Token Prediction (UNTP)}
Text2Token~\cite{an2025text2token} is inspired by empirical findings \cite{nie2024text}: when obtaining a text embedding from the LLM-based embedder, the tokens with the highest decoding probability are the key tokens of the input text. Therefore, Text2Token explores whether the opposite approach holds: if leveraging a given target token distribution as the supervisory signal, the embedders can learning the embedding similar to contrastive fine-tuning. Using data-based statistical methods or model-based filtering methods, Text2Token constructs the target distributions for key tokens in an unsupervised manner and achieves performance comparable to unsupervised LLM2Vec~\cite{behnamghader2024llm2vec}.

\paragraph{Supervised Next Token Prediction (SNTP)} 
Inbedder~\cite{peng2024answer} explores training models using query-answer pairs to enable the LLM-based embedders to follow instructions. The authors argue that for specific questions about textual content, the consistency of LLMs' responses reflects the semantic similarity between texts on the subject being inquired. At the same time, to ensure LLMs provide immediate and relevant responses to questions, it is necessary to guide them in generating short yet informative replies. Given these considerations, the author employed plenty of short-answer question-answer pairs to train LLMs and empirically demonstrated that last pooling (corresponding to the generation of the first word) yields the best embedding performance.

\subsubsection{Reinforcement Learning (RL)}
Reinforcement Learning from Human Feedback (RLHF)~\cite{ouyang2022training} serves as the final step in current advanced LLM training pipelines, designed to align with human preferences. The reinforcement learning algorithms within RLHF have evolved from the earliest Proximal Policy Optimization (PPO)~\cite{schulman2017proximal} to methods such as Direct Preference Optimization (DPO)~\cite{rafailov2023direct} and Group Relative Policy Optimization (GRPO)~\cite{shao2024deepseekmath}. Among these algorithms, GRPO dispenses with the need for a value model in PPO and high-quality preference data in DPO, making it the primary technical approach currently explored within the text embedding community~\cite{gui2025search, sun2025grace}. 

The key to GRPO lies in sampling a group of candidate solutions for each sample and conducting group evaluations to estimates the group-relative advantage of each solution. Therefore, unlike previous methods, RL-based approaches require randomness in the embedding process to achieve sampling. The most intuitive approach is therefore to incorporate the generated content into the pooling strategy. The core difference of these approaches lies in designing a reward function, and We highlight the distinctions in reward design as below.

Search-R3~\cite{gui2025search} focuses on retrieval tasks, with its reward simultaneously considering (1) format correctness and (2) the Discounted Cumulative Gain (DCG) metric. The first term ensures the embeddings' accessibility: the text embeddings is originated from a special token generated after the reasoning content, thus the reward remains non-zero only when the generated content contains this special token. And the second term directly uses scaled DCG as the reward metric—a straightforward approach. However, since the embedding space continuously evolves during training, re-encoding the entire corpus is impractical. Search-R3 addresses this by introducing a specialized RL environment that incorporates sample forms, document-simplified embeddings, and localized graph refreshes. This environment updates embeddings in only a portion of the corpus at each iteration, mitigating computational overhead.

GRACE~\cite{sun2025grace} focuses on general embedding, and its rewards draw upon the historical successes of (1) contrastive learning and (2) hard-to-learn example mining, while also incorporating the (3) consistency of GPRO sampling. Specifically, the first term emphasizes that the similarity between anchors and positive examples should be higher than that with negative examples; the second term encourages minimizing the similarity to hard negative examples from other samples; while the third term constrains the consistency among embeddings obtained through multiple sampling solutions. Note that the latter two items do not require additional annotation. For the first item, GRACE provides an unsupervised version that encourages the embedding similarity between the two anchor's sampling solutions to be as high as possible. Interestingly, the embedder trained through this algorithm did not lose its generation capabilities.

\subsubsection{Multi-Task Learning (A+B+$\cdots$)}\label{sec:multi-task}

Beyond the standard contrastive learning objective, recent research explores the integration of auxiliary learning objectives during supervised training to enhance representation quality and preserve generative capacities in LLMs. This multi-objective approach aims to broaden the applicability of LLMs across diverse downstream tasks.

\paragraph{Hybrid Loss for Different Data Format}
Contrastive learning, which merely classifies samples as positive or negative examples, often loses finer-grained information such as similarity scores of STS datasets. Advanced models, however, require the full utilization of annotations across various tasks to achieve optimal performance~\cite{huang2024piccolo2}. To fully leverage the fine-grained annotations from the STS task, CoDiEmb~\cite{zhang2025codiemb} simultaneously introduced a Pearson Loss~\cite{zhang2024pcc} to fit similarity scores, a newly proposed Rank KL-divergence Loss to fit similarity rankings, and a RL algorithm called Preference Rank Optimization (PRO)~\cite{peng2024large} to penalize text pairs with reversed rankings. QZhou-Embedding \cite{yu2025qzhou} uses CoSENT Loss \cite{huang2024cosent} to reinforce the ranking relationship between sample similarity scores. Search-R3~\cite{gui2025search} introduces Triplet Loss~\cite{weinberger2009distance} to widen the similarity gap between anchor-positive and anchor-negative pairs.

\paragraph{Contrastive Learning + Generative Objective}
Relying solely on contrastive learning can cause LLMs to lose nearly all their generative capabilities~\cite{muennighoff2024generative}. To enable a model to possess both generative and embedding capabilities simultaneously, or to enhance embedding capabilities by generating additional reasoning steps, it is necessary to find ways to preserve the generative capabilities of LLMs during training. A straightforward approach is to combine generative tasks with contrastive learning for joint training. For example, GritLM~\cite{muennighoff2024generative} demonstrates that simultaneous training with contrastive learning and Next Token Prediction (NTP) enables LLMs to achieve robust text representation while maintaining their native generative abilities. This dual-objective approach balances representational and generative demands. ChatRetriever~\cite{mao2024chatretriever} further refines LLMs for retrieval by integrating contrastive learning with masked instruction tuning on high-quality conversational data, thereby enhancing complex session understanding. O1-Embedder~\cite{yan2025o1} utilizes a single LLM for both query expansion and text embedding. The query expansion is modeled as a next token prediction task, where a large-scale LLM is used to generate the supervised queries. At inference time, the embedding of the original query and O1-Embedder's newly generated query are pooled equally as the final query embedding. ULLME~\cite{man2024ullme} proposes Generation-Representation Learning (GRL), a fine-tuning technique that jointly optimizes contrastive learning and generation objectives. GRL minimizes the Kullback-Leibler (KL) divergence between similarity scores derived from learned representations and the generation probability distributions, ensuring internal consistency between the representational and generative aspects of the model. 

\paragraph{Contrastive Learning + Self-Distillation} DEBATER~\cite{ji2025learning} introduces a continuous thinking process to augment dense retrieval, which generates a special token sequence representing the chain of thought before producing the final document embedding. During training, the maximum similarity score between any token in the chain and the query then determines the final similarity. And an additional self-distillation term that preserves the use of only the last token's embedding can achieve rankings as close as possible to those from all tokens.  Therefore, DEBATER can use only the embedding of the last token during inference, reducing computational complexity.

\paragraph{Contrastive Learning + Regularization} Mistral-SPLADE \cite{doshi2024mistral} use the FLOPs regularization \cite{paria20220minimizing}, applying to the representations in the vocabulary space to ensure the desired sparsity factor. GIRCSE~\cite{tsai2025let} leverages the generative capabilities of LLMs to produce multiple soft tokens, each weighted by multiple token embeddings. The last hidden state of all soft tokens is averaged to serve as the embedding for the entire text. Beyond contrastive learning, the authors introduce Iterative Contrastive Refinement (ICR) to constrain the contrastive loss, ensuring that the loss calculated from the average pooling of the $K+1$ soft tokens is lower than that calculated from the $K$ soft tokens.

\subsubsection{Multi-Stage Training (A→B→$\cdots$)}\label{sec:multi-stage}

\paragraph{Pretext Task $\rightarrow$ Contrastive Learning}   
Neither BERT nor current LLMs are primarily designed as embedding models during their pre-training phases. Consequently, considerable research focuses on adapting the base pre-trained language models before contrastive learning to make them more suitable for embedding tasks. During the BERT pre-training era, representative works include CoCondenser~\cite{gao-callan-2022-unsupervised} and RetroMAE~\cite{xiao-etal-2022-retromae}, which have designed distinct training objectives to enhance BERT's text embedding capabilities. Shifting to the era of LLMs, the lack of semantic aggregation capability in language models remains unchanged; therefore, similar approaches have been extended to improve LLMs' embedding abilities. For the unsupervised setting, Llama2Vec~\cite{li2024llama2vec} employs various templates that enable the model’s hidden states to predict which words from the original sentence and the subsequent sentence are included. Correspondingly, it utilizes two auxiliary pre-training tasks, EBAE (Embedding-Based Auto-Encoding) and EBAR (Embedding-Based Auto-Regression). These two tasks serve as a secondary pre-training technique to enhance the model's ability to capture global semantics using the hidden state from a given position. For the supervised setting, Anchor \cite{su2025training} introduces the bi-directional reconstruction for query-document pairs. Specifically, the authors introduce Embedding-Based Query-to-Document (EBQ2D) to reconstruct document text from query embedding and Embedding-Based Document-to-Query (EBD2Q) to reconstruct query text from document embeddings. Following the long text compression method, AutoCompressor~\cite{chevalier2023adapting}, AutoRegEmbed~\cite{deng2025following}, append multiple special tokens to the end of the query text, and the document is predicted by extracting the last hidden state corresponding to these special tokens. Some pretext tasks are highly relevant to the design of other components, such as mitigating the impact of bidirectional attention switching on LLMs~\cite{behnamghader2024llm2vec} or ensuring the embedder maintains format correctness during inference~\cite{gui2025search}. These methods have been introduced in detail in other sections, so we will not elaborate further here.

\paragraph{Multi-Stage Contrastive Learning}
During the supervised learning phase, numerous studies have explored multi-stage contrastive learning training methods to enhance further the generalization and versatility of the final embedding models. For instance, influenced by earlier work from the BERT era, models like GTR~\cite{ni2022large}, GTE-Qwen2~\cite{li2023towards}, and BMRetriever~\cite{xu2024bmretriever} have investigated two-stage training strategies that combine weakly supervised contrastive learning (WCL) with supervised contrastive learning (SCL). Weakly supervised contrastive learning primarily leverages a large amount of weakly supervised relevance data collected from public domains (e.g., neighboring text spans and question-answer pairs from the QA community) for training. Although this data may contain noise, the extensive domain coverage, typically comprising over a billion data points, significantly improves the model's domain generalization capabilities. Additionally, NV-Retriever proposes a phased supervised training approach tailored to different tasks (e.g., retrieval and STS tasks). Specifically, NV-Retriever \cite{moreira2024nv} utilizes only retrieval-supervised data with in-batch negatives, alongside mined hard negatives, for the first stage, while blending data for retrieval tasks with datasets from other tasks for the second stage.

\subsubsection{Model Merging (MG)}

Model Merging~\cite{izmailov2018averaging,wortsman2022model} was initially proposed to average multiple checkpoints fine-tuned on the same task, enhancing the model's generalization performance. Following model soups approach~\cite{wortsman2022model} in this line, Gemini Embedding~\cite{lee2025gemini} weights multiple final checkpoints obtained from different runs to enhance embedding performance.  Subsequently, the merging of general models and fine-tuned models was demonstrated to achieve superior zero-shot performance~\cite{wortsman2022robust}. LM-Cocktail~\cite{xiao2024lm} pioneered testing on PLM-based embedders, weighing the parameters of the general embedder against those of the fine-tuned models on target tasks. The embedder with the weighted parameter can simultaneously preserve the target task's performance while maintaining the general capabilities across other tasks. With the introduction of methods such as task arithmetic~\cite{ilharco2023editing} and DARE~\cite{yu2024language}, this technique was applied to model checkpoints fine-tuned for different tasks on the same starting point (identical pre-trained models). In the field of text embedding, different tasks do not necessarily reinforce each other; in fact, they can even conflict. For instance, Cpt~\cite{neelakantan2022text} observes that retrieval/classification performance improved while STS performance declined as the number of training steps increased; InstructOR~\cite{su2023one} observes that training symmetric and asymmetric tasks together without adding different instructions leads to a decline in performance. These issues were initially overlooked, then partially addressed by the methods based on instruction tuning. However, the data size across different tasks can vary dramatically. In traditional multi-task learning, this issue can only be addressed through technically challenging methods, such as task gradient weighting~\cite{chen2018gradnorm} or resampling~\cite{arivazhagan2019massively}. As a post-processing technique, model merging is significantly less difficult to implement than multi-task learning, and its performance has progressively reached and even surpassed that of the latter during its development \cite{choi2024revisiting}. Therefore, some empirical studies have applied task-based arithmetic and its variants to general embedding~\cite{li2024improving} and dense retrieval~\cite{braga2025investigating, sasaki2025effect}. In the implementation of the advanced embedder, Lychee~\cite{zhang2025phased} divides all training data into four categories: (1) basic relevance retrieval, (2) code retrieval, (3) tool retrieval, and (4) complex instruction-based retrieval, fine-tuning one embedder for each category dataset from the same initiation point. Then, the merging method called spherical linear interpolation (SLERP)\footnote{\url{https://github.com/Digitous/LLM-SLERP-Merge}} is introduced to merge four embedders with an efficient hyper-parameter search. Note that Qwen3 Embedding~\cite{zhang2025qwen3} also utilizes SLERP, but the method for acquiring the merged checkpoints remains unclear. Recently, CoDiEmb~\cite{zhang2025codiemb} A first employed the model soups approach to fuse separate models for the IR and STS tasks. Subsequently, it applied finer-grained layer-wise weights to adjust the soup models of IR and STS, ultimately yielding the final checkpoint.

\subsection{Commercial Service}

Many commercial companies provide generic text embedding services to support traditional NLP tasks and RAG. In English text embedding service, it mainly includes OpenAI \footnote{\url{https://openai.com/index/introducing-text-and-code-embeddings}}, Google \footnote{\url{https://cloud.google.com/vertex-ai/generative-ai/docs/embeddings/get-text-embeddings}}, Amazon \footnote{\url{https://docs.aws.amazon.com/bedrock/latest/userguide/titan-embedding-models.html}}, Alibaba Cloud\footnote{\url{https://www.alibabacloud.com/help/en/model-studio/developer-reference/general-text-embedding}}, ByteDance\footnote{\url{https://seed1-5-embedding.github.io/}}, Voyage \footnote{\url{https://docs.voyageai.com/docs/embeddings}}, Cohere \footnote{\url{https://cohere.com/blog/introducing-embed-v3}} and Jina \footnote{\url{https://jina.ai/embeddings}}. Due to the trade secrets involved, only part of companies have revealed some of the core technology behind their services. For example, OpenAI simultaneously provides text embeddings with different dimensional settings for different levels of time-storage sensitivity in practice, where matryoshka representation learning (MRL) \cite{kusupati2022matryoshka} is considered to be the key to obtaining the embedding with flexible dimensions using only one encoder. Google demonstrates that Gecko \cite{lee2024gecko}, which achieves superior performance on 1B scale and 786 dimensions, relies heavily on a two-stage synthetic data generation process with LLMs: the first step generates diverse tasks and queries based on the instruction and the second step generates positive and negative samples based on the obtained tasks and queries. As shown in Table \ref{tab:embedders}, many commercial companies publish their own technical reports. However, with the exception of a few that are fully open-source~\cite{zhang2025f2llm}, most do not release their source code or data. Furthermore, they do not disclose all details regarding their data processing or training methods. Some companies (such as Bytedance and Voyage) have published results on public benchmarks, but only provided API services without further details. In addition, there have been the attempt~\cite{tamber2024can} that attempt to distill knowledge through embeddings obtained from commercial APIs, successfully obtaining a local embedder with similar performance.

\section{Text Embedding Understanding with LLMs}
\label{sec:embedding_understand}

Large language models have a powerful paraphrase capability that can be quickly aligned with and interpreted in a variety of already learned embedding spaces of image \cite{merullo2023linearly}, item \cite{tennenholtz2024demystifying,lei2024recexplainer} and concept \cite{teehan2024college}. Thus, for different purposes, existing work attempts to make an understanding of text embeddings with the help of LLMs. There are two embedding-related tasks in NLP: long context compression (ICC) and embedding inversion (EI).

\subsection{Long Context Compression}
\label{sec:method_lcc}
Long context compression proposes that the input for conditioning a language model can be condensed into a smaller, specifically chosen set of words or dense vectors. This reduces context length, leading to faster inference speeds and smaller storage requirements.
Here, we distinguish long text compression from two similar research directions: (1) language models with memory mechanisms \cite{liu2018generating,dai2019transformer,rae2019compressive,wu2021memorizing,zhang2021poolingformer,munkhdalai2024leave,bulatov2022recurrent} and (2) key-value compression. The language model with memory mechanisms is designed to improve the in-context length of language modeling, and key-value compression is a technique proposed to improve the decoding efficiency of LLM based on key-value.

First, we can distinguish between long context compression and language modeling with memory mechanisms in design principle: Language modeling with memory mechanisms is for the native architectural design, and the memory module is part of the language model and requires full-parameter pre-training so that the model learns to read (comprehension) and write (update) the memory. In contrast, Long text compression is a post-adaptation of existing LLMs, which does not or slightly changes the parameters and architecture of the learned LLMs. The key to this task is to learn a compressor that can be aligned with the LLM input space.

Next, we can distinguish between long text compression and key-value compression in terms of the form of the input. Key-value compression aims to design an algorithm for only pruning the key-value cache, whose inputs are generated by the attention mechanism in LLM. On the contrary, the input of long text compression is text, although its output may be key-value embedding.

In addition, the methods of long text compression are usually categorized into two main groups based on the output form : (1) dense embeddings and (2) discrete tokens. We introduce the methods whose output are dense embeddings in Section \ref{sec:soft_prompt} and focus on the methods whose output are discrete tokens in Section \ref{sec:context_dist}.

\subsubsection{Soft Prompt}\label{sec:soft_prompt}
Prompt tuning \cite{lester2021power} introduces the trainable, parameterized soft prompt to adapt pre-trained language models to specific downstream tasks in natural language processing. In prompt compression, the general idea of this method is to align the soft prompt with the representation of the original long text, thereby enabling the replacement of the original long text with the soft prompt in practical applications. Different methods propose different optimization objectives to obtain sufficiently short soft prompts that can preserve the semantic meaning of the original text as much as possible. According to the training tasks, these methods can be divided into three categories: alignment, prediction, and restoration.

\paragraph{Alignment-Based Methods} The alignment method uses Kullback-Leibler (KL) divergence as the optimization objective. PC~\cite{wingate2022prompt} trains a small set of adaptable soft prompt weights to closely mimic the distribution induced by a larger, fixed hard prompt using KL divergence, resulting in a compact representation that retains essential information for guiding language model generation. Gist~\cite{mu2023learning} adopts a special attention mask strategy in training to align gist tokens and task prompts. HD-Gist~\cite{jiang2024hierarchical} creates hierarchical and dynamic gist tokens based on Gist for tool usage scenarios. Gist-COCO~\cite{li2024say} also uses KL divergence to align the gist representations obtained from the encoder with the original context. QGC~\cite{cao2024retaining} uses queries to guide the context compression process and align the output by the LLM with the ground-truth answer using KL divergence.

\paragraph{Prediction-Based Methods} The prediction method takes the language model task of predicting the next word as the optimization objective. \cite{chevalier2023adapting} adapt LLMs into \textit{AutoCompressors} by compressing long contexts into summary vectors. During training, summary vectors are trained using an unsupervised approach, where long documents are segmented and summary vectors from previous segments are incorporated into language modeling. All summary vectors are concatenated during inference to form the soft prompt for the long text.~\cite{lin2021readonce} extend standard text-to-text transformer models to representation+text-to-text models by transferring contexts to shorter embeddings.~\cite{cheng2024xrag} adopts a two-stage training strategy to enable a text encoder to compress a long context into a single input token embedding for an LLM.~\cite{shao2024flexibly} adopts an LLM to transform the token embedding of contexts to shorter embeddings.~\cite{gao2024unifying} combines contrastive learning and language model loss to compress and select demonstrations simultaneously.

\paragraph{Restoration-Based Methods} The restoration method uses text reconstruction in autoencoders as the optimization objective. ~\cite{ge2023context} utilizes an encoder-decoder structure and trains \textit{memory tokens} in an reconstruction manner. The original context is converted into shorter \textit{memory token} embeddings, which are then used to reconstruct the context.~\cite{wang2024context} employs a set of learnable digest embeddings to condense contextual information, producing digest vectors.~\cite{huang2024recurrent} utilizes an encoder to segment long contexts into fixed-length embeddings, which are then input into the LLM along with prompts. Then, an instruction reconstruction task is proposed to rebuild the instructions and generate answers.~\cite{xu2024concise} retains key information and compresses other contents into shorter soft prompts. PE-RANK~\cite{liu2024leveraging} compresses the embedding of each passage into a passage representation. It employs a two-stage ranking training method: in the first stage, reconstruction loss is used to enable the LLM to reconstruct the corresponding paragraph from the input embeddings; in the second stage, ranking loss is applied to equip the LLM with ranking capabilities. \cite{deng2024silver} proposed fine-grained autoencoding and segment-wise token importance estimation to enhance gist-based context compression.

\subsubsection{Context Distillation}\label{sec:context_dist}
This method removes unimportant words from the context, retaining only the words crucial for the language model, thereby compressing the context length at the input level. There are primarily two methods: the first transforms long contexts into sequences of important tokens, while the second converts long contexts into natural language summaries or context snippets. These methods can be divided into selective and generative.

\paragraph{Extraction-Based Methods} The selective method relies on the capabilities of the LLM itself to directly score each token.~\cite{jiang2023llmlingua} employs a small language model to iteratively select the most important tokens by calculating perplexity.~\cite{li2023compressing} also utilizes a small language model and computes self-information for each token. Nugget~\cite{qin2023nugget} proposes a text representation method for an encoder-decoder structure that maps a sequence of tokens to a shorter one by utilizing a trained scorer to select top-k tokens in the last layer of the encoder. Nugget2D~\cite{qin2023nugget} extends \textit{Nugget} to the decoder-only LLMs like LLaMA by utilizing selected tokens in all layers of transformers rather than only in the last layer. This method uses a small number of tokens to represent the semantic information of context and generally does not require additional annotated data for training.
However, this method cannot produce complete natural language paragraphs that humans can understand, leading to a lack of interpretability.~\cite{jung2023discrete} utilizes supervisory signal generated using by LLM to optimize a language model scoring system through reinforcement learning.~\cite{berchansky2023optimizing} removes tokens with low cross-attention scores within the context of two documents.

\paragraph{Abstract-Based Methods} The generative method relies on appropriate text summarization techniques.~\cite{chuang2024learning} trains a small Language Model (LLM) by aligning compressed context and original context embeddings while exploiting language model loss using compressed context in specific tasks. It ensures both the natural language format and the consistent utility of the original context.~\cite{xu2023recomp} exploits the text summarization technique to transform the original context into shorter and more coherent natural language summaries.~\cite{fei2023extending} uses a pre-trained embedding model to extract sentence features and cluster similar sentences into text blocks. Then, the text summarization technique is used to obtain the final context.~\cite{wang2023learning} scores each text block in the context and extracts context snippets as the compressed prompt. The advantage of this method is that it can produce natural language paragraphs that are understandable to humans, but it also comes with higher training costs.~\cite{ali2024prompt} leverages the prompt's textual information to construct a graph, then utilizes a graph encoder to extract important elements from the graph to generate a summary.

\begin{table*}
    \centering
    \caption{The Overview of Text Embedding Understanding With LLMs. The {\bf paradigm} part is shortened as follows: Language Model (LM), Context Distillation (CD), Auto Encoding (AE), Context Generation (CG) and Progressive Context Generation (PCG). The {\bf task} part is shortened as follows: Prompt Compression (PC), In-Context Learning (ICL), Retrieval-Augmented Generation (RAG), Information Retrieval (IR), Attribute Inference (AI), and Embedding Inversion (EI).}
    \resizebox{\textwidth}{!}{
    \begin{tabular}{lllllllllll}
    \toprule
    \multirow{2}{*}{\bf Method} & \multicolumn{3}{c}{\bf Model} & \multicolumn{2}{c}{\bf Training Data} & \multicolumn{2}{c}{\bf Training Method} & \multicolumn{2}{c}{\bf Evaluation} \\
    \cmidrule(r){2-4} \cmidrule(r){5-6} \cmidrule(r){7-8} \cmidrule(r){9-10}
    & Embedder & Projector & LLM & Scale & Form & Paradigm & Projector & CR & Task \\
    \midrule
    GEIA \cite{li2023sentence} & SRoBERTa, SimCSE, ST5, MPNet & GPT-2 & 133K-209K & X & CG & & AI, EI \\
    Vec2Text \cite{morris2023text} & GTR, OpenAI API & T5 & 5M & X & PCG & 1-Layer MLP & EI \\
    M-Vec2Text \cite{chen2024text} & T5 & ME5 & 3M-5M & X & PCG & 1-Layer MLP & EI \\
    qsT5 \cite{adolphs2022decoding} & GTR & T5 & 3M & (I, X, Y) & PCG & 1-Layer MLP & IR \\
    Text Revealer \cite{zhang2022text} & BERT, Tiny-BERT & GPT-2 & - & X & CG & - & EI \\
    Tranfer Attack \cite{huang2024transferable} & OpenAI API, SBERT, ST5 & GPT-base & - & X & PCG & 1-Layer MLP & EI \\
    \bottomrule
    \end{tabular}}
\end{table*}

\subsection{Method for Embedding Inversion}
\label{sec:method_inversion}
In this section, we consider the privacy leak problem from text embedding. Models for text embedding aims to train universal vector representations, these embedding can support many downstream tasks. There are many models that serve the text embedding purpose, such as BERT~\cite{kenton2019bert}, Sentence-T5~\cite{ni2022sentence}, GPT-2~\cite{radford2019language}, GTR~\cite{ni2022large}. With the widespread use, the issue of privacy leakage from text embedding has become increasingly important. We introduce this issue from the perspective of embedding attacks. Depending on the target of the attack (or the content of the leakage), they are mainly divided into attribute inference attack, embedding inversion attack, and model inversion attack~\cite{li2023privacy}. Here, we focus on attacks related to large language models.

\paragraph{Attribute Inference Attack} The attacker attempts to deduce some of the original text's information, such as keywords, phone numbers, ID card numbers, etc., from direct text embeddings or gradients of embedding models. So at this point, the leaked information consists of some words from the original text, it's word-level privacy leakage. \cite{pan2020privacy} used GPT-2~\cite{radford2019language} as one of the target models to embed airline reviews and public healthcare records, then constructed a multi-class classification model to infer attribute words (passenger's residence and itinerary; patient's disease type, etc.) from text embedding. The experimental results indicate that GPT-2 is more susceptible to privacy leakage than other embedding models. Additionally, attribute information can also be inferred from gradients in the federated learning of large language models for text embedding. \cite{balunovic2022lamp} leverages GPT-2 \cite{radford2019language} to enhance the search for natural text and employs a mix of continuous and discrete optimizations to minimize loss and escape local minima; it successfully reconstructs original text from gradients. \cite{fowl2022decepticons} exploit the Transformer architecture and token embedding of GPT-2 \cite{radford2019language} and attack the embedding model by deploying malicious parameter vectors to reveal some tokens of private user text. ~\cite{gu2023towards} used a decoder-only attack model to decode a sentence embedding generated by GPT-2~\cite{radford2019language} 10 times to collect potential outputs, filter out stopwords, sort the remaining words by frequency, and then select the top-k words for examination. However, in reality, attribute inversion can not only be used for attacks but also for beneficial purposes. \cite{avitan2024naturallanguagecounterfactualsrepresentation} converted representation counterfactuals into string counterfactuals, offering a deeper interpretability of the model's feature encoding for particular concepts. By using this method, on one hand, it is possible to modify the privacy data in the representation space for privacy protection, and on the other hand, it can correct the biases in classification through data augmentation.

\paragraph{Embedding Inversion Attack} An embedding inversion attack aims to reconstruct all the text from the corresponding embeddings, not just a collection of words. So, the leaked information is the original text; it's sentence-level privacy leakage. \cite{adolphs2022decoding} focused on the retrieval framework, which trained a pseudo-relevance feedback (PRF) T5 model~\cite{raffel2020exploring} as query decoder to decode query text from embeddings, which GTR~\cite{ni2022large} model generates, and this kind of decoder can be used to generate new related queries. ~\cite{morris2023text} considered the embedding inversion problem as controlled generation, and proposed Vec2Text method to reconstruct the full text represented in dense text embeddings encoded by GTR-base~\cite{ni2022large} and text-embeddings-ada-002 available via the OpenAI API. Vec2Text guesses an initial text and iteratively refines this text by re-embedding and correcting it, with T5~\cite{raffel2020exploring} as decoder module. ~\cite{chen2024text} extended Vec2Text with Ad hoc Translation and Masking Defense Mechanism, so that this method can be used in multilingual and cross-lingual scenarios.~\cite{zhuang2024understanding} made deeper understanding of Vec2Text methods, and mitigates the risk of text recoverability using a fix for emmbedding transformation. ~\cite{li2023sentence} aimed to recover the target sequences word by word directly using generative decodes. To reconstruct input text from its embeddings from various models such as Sentence-BERT~\cite{reimers2019sentence}, SimCSE~\cite{gao2021simcse}, Sentence-T5~\cite{ni2022sentence}, they proposed a generative embedding inversion attack (GEIA) in which a GPT-2~\cite{radford2019language} is trained as the attacker model.

\paragraph{Model Inversion Attack} Unlike attribute inference attacks and embedding inversion attacks, the target of model inversion attacks is not the input text but the text embedding model itself. \cite{zhang2022text} proposed Text Revealer as a model inversion attack for text reconstruction against text classification with transformers. Based on an external dataset and the black-box access to the original text embedding model, it constructed an (embedding, text) training dataset, which is used to train a GPT-2~\cite{radford2019language} as a text generator. Using this text generator and perturbing the hidden state optimally with the feedback of the target model, the text revealer could reconstruct private texts in the original training data. ~\cite{huang2024transferable} proposes a transfer attack method that employs a surrogate model to emulate the victim model, aiming to steal the text encoder from the victim model. This paper also uses the GPT-base model as an embedding-to-text part to invert embeddings to their original text sequences.

\section{Challenge}
\label{sec:challenge}
\subsection{Existing challenges}

\subsubsection{False Negative detection}

Under the embedding training framework of contrastive learning, the model aims to learn appropriate embeddings by pulling positive sample pairs closer while pushing negative sample pairs farther apart. The appearance of false negatives disrupts this learning process. Samples that initially belonged to the same category (positive sample relationship) are wrongly regarded as negative samples to participate in the training. This causes the model to receive incorrect "supervisory signals", resulting in the learned embeddings failing to accurately reflect the real semantic relationships among samples. An intuitive idea to solve the false negative problem is to correct the wrongly labeled samples through accurate annotations. However, whether it is the traditional manual annotation method or annotation with the help of large-scale language models (LLMs), both face the challenge of high costs. For manual annotation, in the face of massive data, it is almost an impossible task to expect to annotate thousands of samples for each query. Manual annotation is not only time-consuming and labor-intensive but also easily affected by subjective factors of annotators and issues such as the consistency of annotation standards. On the other hand, although LLMs have demonstrated powerful capabilities in many fields such as natural language processing, using them for annotation also comes with relatively high cost expenditures, including the cost of calling APIs, possible errors in generated results, and limitations on the accuracy of annotating domain-specific knowledge. Therefore, relying on large-scale and accurate annotations to eliminate false negatives has significant obstacles in reality.

To address the false negative problem and the challenge of annotation costs, some studies have proposed methods such as improving the loss function or negative sample sampling strategies to attempt to alleviate this problem. For example, the peer loss method~\cite{wang2024mitigating} designs a regularization term for the loss function, aiming to reduce the adverse impact of false negatives on the overall training gradient. ~\cite{zhou2022simans} believes that samples that are similar to positive samples but not similar to the query should be sampled, and such samples are more likely to be hard negatives rather than false negatives. However, these methods still have obvious limitations. On the one hand, they often rely heavily on the practical experience of researchers and lack a solid theoretical basis as a guarantee. This means that in different datasets and task scenarios, it is difficult to maintain the effectiveness of the methods stably, and there are doubts about their generalization ability. On the other hand, many improved loss functions have only been experimentally verified on a small range of data, and it cannot be proved whether they can truly and completely solve the false negative problem in a large-scale and diversified data environment. When applied to more complex and larger-scale practical tasks, these methods may not achieve the expected results, and the model training will still be interfered by false negatives, resulting in a significant reduction in the quality of embeddings and the performance of subsequent applications.

In conclusion, the false negative problem is a key issue that urgently needs to be further conquered in the process of embedding training based on contrastive learning. Although existing solutions have made some attempts, there is still a large gap from completely and effectively solving it. In the future, more in-depth theoretical research and more universal and scalable solutions are needed to meet this challenge, thereby promoting the quality and effect of embedding training to a new level.

\subsubsection{The Curse of Low-Resource Languages}
The development of multilingual text embedding is heavily influenced by the development of large language models. On the one hand, advanced LLMs effectively improve the quality of text embedding through data augmentation \cite{wang2024improving}; on the other hand, LLM-based embedders will perform much better in seen languages than in unseen languages during pre-training \cite{zhang2023language}. However, because those really low-resource languages are unable to provide enough high-quality corpus, they have difficulty getting support for advanced LLMs. Obviously, this situation has become a vicious circle. From the original point of view, there are more than 7000 languages in the world, and the Unicode-based tokenizers only support 168 of them \footnote{Unicode 16.0, which is released in September 2024, contains 168 scripts.}, which means that the vast majority of languages cannot be theoretically supported by the LLMs. Even for those languages in Unicode, the unfairness of the training corpus size, quality, and token distribution leads to huge differences in decoding performance \cite{ali2024tokenizer,ansell2023unifying} and efficiency \cite{limisiewicz2024myte,ahia2023all} across languages.

\subsubsection{Native High-Quality Embedding}

It has to be admitted that the text embedding is not good enough for both traditional PLMs and advanced LLMs, so they perform poorly on STS, IR, and other tasks in the zero-shot setting. However, the reasons behind this phenomenon are not yet fully understood. In some works, anisotropy of the embedding space is regarded as the reason for the current problem. \cite{ethayarajh2019contextual} firstly find that the token embedding space in PLMs, including BERT, ELMo, and GPT-2 is anisotropic. It results in token embeddings in a narrow, high-dimensional conical space, with no obvious correlation between word and word similarity and semantics \cite{li2020sentence}. The embeddings of sentences and paragraphs are usually obtained from their token embeddings with a pooling strategy, sharing the same space with the token. Therefore, the anisotropy space leads to unrelated texts with high similarity. Note that there is some controversy to the conclusion here, including (1) The degree of anisotropy should not be measured using the cosine similarity \cite{rudman2022isoscore,cai2021isotropy}; (2) Anisotropy may not seriously harm downstream task performance and even be profitable \cite{ait2023anisotropy,rudman2023stable}; (3) Anisotropy is not inherent to Transformers \cite{machina2024anisotropy}.
Some efforts were to improve isotropy in the embedding space, such as post-processing methods \cite{li2020sentence,su2021whitening,rajaee2021cluster}, contrastive learning \cite{gao2021simcse,su2022contrastive}, and other regularization items \cite{gao2019representation,zhang2020revisiting}. However, these methods evaluate performance on either the generation task or downstream tasks, and it is not clear that any method can perform well on both text generation and text embedding simultaneously. In the era of PLMs, almost all downstream tasks need to be fine-tuned on top of PLMs, so fine-tuning with contrastive learning to obtain good text embeddings is a no-bad approach. In the era of LLMs, LLMs can accomplish many tasks directly using generative paradigms. At this point, it seems like putting the cart before the horse compromises the generative capabilities of the larger model to obtain high-quality textual embeddings.

\subsection{New challenges}

\subsubsection{Complex Instruction Following}
In recent times, numerous embedding approaches based on language models (LLMs) claim to have trained with instruction tuning approach. The aim of such training is to improve the models' ability to understand and perform specific tasks according to the provided instructions. However, despite these claims, current instruction models encounter significant difficulties when trying to effectively meet complex retrieval demands. Many researchers are attempting to develop new benchmarks to evaluate these models' capabilities in following complex instructions. For example, tasks involving the precise chronological order of events~\cite{xiao2024rar}, or those related to elaborate narratives, such as creating a detailed summary of a long fictional story or extracting key plot elements in a specific order from a complex literary work~\cite{weller2024followir}. Additionally, there are reasoning-based retrieval tasks, like finding the solution code for a LeetCode problem~\cite{su2024bright}. The current inability of instruction models to handle these complex retrieval situations well indicates another crucial research area. This area undoubtedly requires the careful attention of the academic and research communities. By concentrating on improving these aspects, it is possible to potentially enhance the overall performance and practical usefulness of these language models in a wide range of applications.

Additionally, although substantial progress has been made in text embedding methods based on Instruction Tuning, it is challenging for users to clearly and accurately describe their intent in practical applications. Therefore, automatically generating clear descriptions based on available information, such as the type of corpus, the user's query, and other background information, represents a direction that requires further exploration.

\subsubsection{Privacy Leakage from Text Embedding}
As embedding technology continues to evolve, it has been realized that the rich information contained in an embedding may be a threat to privacy security. \cite{pan2020privacy,song2020information}. In the era of LLMs, retrieval-augmented generation \cite{lewis2020retrieval} has become a necessary technique in many LLM-based services. In addition, the need for larger-scale data and knowledge has led to commercial services like vector databases and text embedding API. These trends have led to an unprecedented focus on information security in text embedding.

As the embedding inversion task demonstrates, the text can be partially or even fully restored from its embedding without accessing the embedders' parameters. While appropriate noise was shown to be resistant to trained decoders \cite{morris2023text}, it was not possible to confirm that the fine-tuning of decoders on noisy data could overcome noise interference.

In fact, we can see the antagonism in the two tasks elaborated in Section \ref{sec:embedding_understand}: The methods in long context compression aim to compress long text losslessly, while in embedding inversion, the works warn that the information embedded in the embedding can be easily restored and try to design some defenses.

\subsubsection{High-Dimensional Text Embedding}

Scaling the model size but fixing the embedding dimension has been shown to be an effective way to improve the encoder's performance \cite{ni2022large}. However, increasing the dimension of hidden states with layer and attention head number has been a generalized method when scaling LLMs' size. Therefore, LLMs' hidden states generally have a dimension over 2,048, which is 2x and 2.67x than that of BERT-Large and BERT-base, separately. When LLMs are used as embedders, the increased embedding dimensions exacerbate the computational and storage overhead.

To achieve a flexible trade-off between performance and efficiency, Matryoshka representation learning (MRL) \cite{kusupati2022matryoshka} uses the special optimization objectives during training that direct the most critical information to the part dimensions of the embedding. This technique has already landed in commercial services \footnote{\url{https://openai.com/index/new-embedding-models-and-api-updates}}. In addition, some post-processing approaches \cite{raunak2019effective,hwang2023embedtextnet,xue2024word} have been proposed to reduce the dimensionality of the learned embeddings. A recent work, CSR \cite{wen2025beyond}, introduces automatic coding and task-aware contrastive learning to learn sparse representations. The authors show the sparse representation can maintain almost the same performance as the original LLM-based Embedder in 32 dimensions, demonstrating the huge potential for post-sparse techniques.

In addition, there are many deeper issues to be explored. For example, (1) what is the theoretical lower bound on the dimensionality of lossless compression; (2) how to design more efficient compression algorithms to obtain embeddings with higher information density; and (3) how to derive sparse representations that are more friendly to long documents. However, dimension, as one of the most important measures of embedding quality, still has many deeper issues to be explored. For example, (1) what is the theoretical lower bound of dimensionality for information lossless compression of data; (2) how to realize the efficient conversion of low information density high-dimensional embeddings to high-information density low-dimensional embeddings; and (3) how to convert text embedding to sparse representations that are more friendly to long documents.

\subsubsection{Training \& Inference Overhead for LLMs}

LLMs usually have more than 1B parameter, which makes full parameter fine-tuning very difficult. Various parameter-efficient fine-tuning methods, such as Adapter \cite{houlsby2019parameter}, Soft Prompt \cite{lester2021power}, and LoRA \cite{hu2021lora}, alleviate the computation and memory overhead to some extent. However, the time overhead is still huge, making fine-grained ablation experiments for hyper-parameters difficult to accomplish. Some work has failed to explore the full potential of the method by training a fixed number of steps \cite{behnamghader2024llm2vec} or epochs \cite{muennighoff2024generative}. In addition, how model compression techniques, including embedding sparsification \cite{nie2024text}, parameter pruning \cite{li20242d}, and knowledge distillation \cite{xiao2022distill}, will affect the embedding quality has not been fully explored yet.

\section{Future Trends}
\label{sec:trends}

\subsection{Methods for Cross-Lingual \& Cross-Modal Domain}
Advanced LLMs and MLLMs already support over a hundred languages \cite{lai2024llms} and can understand the information in multiple modalities such as vision \cite{liu2024visual}, audio \cite{lyu2023macaw}, and even physical fields \cite{chen2024building}. The powerful understanding of LLMs and MLLMs has been leveraged to represent multi-lingual and multi-modal data.

\subsubsection{Cross-Lingual Text Embedding}
In cross-lingual scenarios, the downstream tasks, such as STS \cite{cer2017semeval,chen2022semeval} and retrieval \cite{zhang2021mr,zhang2023miracl}, have sufficient high-quality datasets for evaluation. With the help of multilingual PLMs models (e.g., mBERT \cite{kenton2019bert} and XLM-R \cite{kenton2019bert}), mDPR \cite{zhang2023toward}, mE5 \cite{wang2024multilingual}, BGE-M3 \cite{chen2024bge}, mGTE \cite{zhang2024mgte} and KaLM-Embedding \cite{hu2025kalm} have achieved better results on various types of downstream tasks.

\paragraph{LLM-Based Embedder} The common backbones used by LLM-based embedders, such as Mistral and Qwen2, are pre-trained on large multi-lingual corpora, which leads us to expect higher-quality cross-language text embeddings using LLMs. UDEVER \cite{zhang2023language} shows the powerful embedding generalization capabilities of multi-lingual LLM, even when fine-tuned using only English text. 
Subsequent embedders based on Mistral or Qwen, such as GTE-Qwen2, GirtLM, and E5-Mistral, have achieved extraordinary performance on MTEB after contrastive learning on large-scale multi-lingual corpora. Specifically, E5-Mistral \cite{wang2024improving} synthesized a large-scale training corpus using LLM to enhance data diversity; GTE-Qwen2 \cite{li2023towards} introduced weakly-supervised-supervised two-stage contrastive learning, which fully utilized the massive noisy web corpus. xVLM2Vec \cite{musacchio2025xvlm2vec} extends cross-modal embeddings from MLLM-based embedders further in cross-language settings through self-knowledge distillation. Gemini-Embedding \cite{lee2025gemini} integrates the experience of synthetic data and two-stage contrastive learning, and based on this, utilizes the model merging technique \cite{wortsman2022model} to improve the embedding performance on multi-lingual tasks further.

\paragraph{Commercial Service} Both OpenAI and Cohere offer multilingual embedding services; Cohere blogged about the second phase of its model training using additional supervised signals generated by LLM \footnote{\url{https://cohere.com/blog/introducing-embed-v3}}; and OpenAI's currently disclosed tech line \cite{neelakantan2022text} and up to 3,072 embedding dimensions give us a reasonable suspicion that the LLM backs the service as an encoder.

\paragraph{Evaluation} 
Cross-lingual embedding as a long-standing research area has diverse downstream task evaluation benchmarks, such as MASSIVE \cite{fitzgerald2023massive} for Classification, STS-17 \cite{cer2017semeval} and STS-22 \cite{chen2022semeval} for semantic text similarity, MKQA \cite{longpre2021mkqa} for question answering, Mr.TyDi \cite{Zhang2021mrtydi}, XOR-Retrieve \cite{asai2021xor}, XTREME-UP \cite{ruder2023xtreme} and MIRACL \cite{zhang2023miracl} for text retrieval, CodeSearchNet \cite{husain2019codesearchnet} for code search. Newly proposed evaluation benchmarks have recently become more difficult or compounded, such as MLDR \cite{chen2024bge} for long document retrieval, mFollowIR \cite{weller2025mfollowir} for instruction following; MTEB(Code)\footnote{\url{http://mteb-leaderboard.hf.space/?benchmark\_name=MTEB\%28Code\%2C+v1\%29}} and MMTEB \cite{enevoldsen2025mmteb} for universal multi-lingual text embedding.

\begin{table*}[t]
    \centering
    \caption{The overview of MLLM-based embedders. The {\bf paradigm} part is shortened as follows: multi-modal supervised contrastive learning (MSCL), text supervised contrastive learning (TSCL), masked next token prediction (MNTP), and next token prediction (NTP).}
    \resizebox{\textwidth}{!}{
    \begin{tabular}{llllllllll}
    \toprule
    \multirow{2}{*}{\bf Method} & {\bf Model} & \multicolumn{4}{c}{\bf Architecture} & \multicolumn{2}{c}{\bf Training} & {\bf Evaluation} \\
    \cmidrule(r){2-2} \cmidrule(r){3-6} \cmidrule(r){7-8} \cmidrule(r){9-9}
     & LLM (Encoder) & Pooling & Attention & Projector & PEFT & Input Format & Paradigm & Task  \\
    \midrule
        E5-V \cite{jiang2024e5} & LLaVa-1.6 & Last & Causal & $\times$ & LoRA & Prompt & MSCL & Flickr30K, COCO, CIRR, FashionIQ, STS  \\ 
        LLM2CLIP \cite{huang2024llm2clip} & \llama+ViT & Mean & Bi & $\surd$ & LoRA & Prompt & MNTP→MSCL & Flickr30K, COCO, ShareGPT4V, Urban-1k, DOCCI  \\ 
        LamRA-Ret \cite{liu2024lamra} & Qwen2-VL & Special Last & Causal & $\times$ & LoRA & Instruction+Prompt & TSCL->MSCL & M-BEIR  \\ 
        DSE \cite{ma2024unifying} & Phi-3-V & Special Last & Casual & $\times$ & LoRA & Prompt & MSCL & NQ, SlideVQA-open  \\ 
        InstructCIR \cite{zhong2024compositional} & LLaVA-Phi-3-mini & Special Last & Causal & $\times$ & LoRA & Instruction & MSCL & CIRCO, CIRR, FashionIQ  \\ 
        VladVA \cite{ouali2024discriminative} & LLaVa-1.5 & Special Last & Causal & $\times$ & LoRA & Prompt & MSCL+NTP & Flickr30K, COCO, SugarCrepe, SugarCrepe++  \\ 
        MMRet-MLLM \cite{zhou2024megapairs} & LLaVa-1.6 & Special Last & Causal & $\times$ & LoRA & Instruction & MSCL & CIR, MMEB  \\ 
        GME \cite{zhang2024gme} & Qwen2-VL & Special Last & Causal & $\times$ & LoRA & Instruction & MSCL & UMRB, BEIR, M-BEIR, ViDoRe  \\ 
        VLM2Vec \cite{jiang2024vlm2vec} & Phi-3.5-V, LLaVa-1.6 & Last & Causal & $\times$ & LoRA / $\times$ & Instruction & MSCL & MMEB  \\ 
        UniSE-MLLM \cite{liu2025any} & Qwen2-VL & Special Last & Causal & $\times$ & LoRA & Instruction & MSCL & MVRB  \\ 
        MM-Embed \cite{lin2024mm} & LLaVa-1.6+NV-Embed & Post & Bi & $\times$ & LoRA & Instruction & MSCL->TSCL & M-BEIR + MTEB  \\ 
        ColPali \cite{faysse2024colpali} & PaliGemma & - (Late Interaction) & Prefix & $\surd$ & LoRA & Instruction & MSCL & ViDoRe  \\ 
        LLaVE \cite{lan2025llave} & LLaVA-OV, AquilaVL & Last & Causal & $\times$ & $\times$ & Instruction & MSCL & MMEB \\
    \bottomrule
    \end{tabular}}
    \label{tab:mllm_embedders}
\end{table*}

\subsubsection{Cross-Modal Text Embedding}
In the cross-modal domain, joint language-vision embedding learning \cite{liu2024universal, lin2023finegrained, wei2023uniir} has garnered the most attention. Its applications have gradually expanded from image-text retrieval \cite{young2014image,lin2014microsoft} to multiple tasks such as composed image retrieval \cite{liu2021image,wu2021fashion}, multi-modal document retrieval \cite{chang2022webqa} and multi-modal knowledge retrieval \cite{luo2023end,hu2023open}, etc. Before the advent of multimodal large language models (MLLMs), ONE-PEACE \cite{wang2023one} raised the number of parameters to 4B based on customizing the architecture, enabling a unified representation of images, text, and audio.

\paragraph{MLLM-Augmented Multi-Modal Embedding} Unlike text data, multimodal data is smaller-scale and noisier, hindering the development of multi-modal Embedding. Recent studies have shown that using LLM to augment multi-modal data while employing good filtering will effectively improve the performance of multi-modal embedding across multiple tasks. For example, VISTA \cite{zhou2024vista} generates high-quality instructions and captions with the powerful OpenAI LLM and Stable Diffusion, achieving substantial improvements on various cross-modal retrieval tasks. SPN \cite{feng2024improving} uses MLLMs and the pre-trained image encoder to synthesize both positive and negative examples for the composed image retrieval task, successfully improving the performance of multiple baseline models. To train the MLLM-based embedder, the larger scale instruction multi-modal dataset, such as MegaPairs \cite{zhou2024megapairs} and VIRA \cite{liu2025any}, are synthesized with LLM-MLLM pipelines. MLLM-Embedder trained on these datasets show strong performance on uni-modal (even visual modal) and multi-modal tasks.

\paragraph{MLLM-Based Embedder} As shown in Table \ref{tab:mllm_embedders}, many attempts at MLLM-based embedders have been inspired by the successful experience of LLM-based embedders. For example, E5-V \cite{jiang2024e5} follow the prompt style of PromptEOL \cite{jiang2023scaling} and fine-tune MLLMs with text contrastive learning only, firstly showing the potential of MLLMs as a multi-modal unified embedder. LLM2CLIP \cite{huang2024llm2clip} trains an LLM-based embedder following LLM2Vec \cite{behnamghader2024llm2vec} and replaces the text encoder of CLIP with this LLM-based embedder, improving the performance of the CLIP model on multiple tasks effectively. MM-Embed \cite{lin2024mm} introduces the incremental text contrastive learning stage after traditional multi-modal contrastive learning, showing strong performance in both text and multi-modal embedding tasks. VladVA \cite{ouali2024discriminative} uses next token prediction to increase the task difficulty for long text, preventing early gradient dissipation in the optimization process. In addition, many effective strategies in text embedding methods are followed, such as instruction tuning \cite{liu2024lamra,jiang2024vlm2vec}, hard negative mining \cite{zhang2024gme, lin2024mm}, scaling negative number \cite{lan2025llave}, etc. Except for MLLM-based embedders for universal multi-modal and uni-modal tasks, some MLLM-based embedders focus on the optimization of complex tasks in real scenarios. For example, DSE \cite{ma2024unifying}, UniSE-MLLM \cite{liu2025any}, and ColPali \cite{faysse2024colpali} focus on tasks related to document screenshot understanding. They leverage MLLM's powerful comprehension capabilities to simplify the original complex embedding process while improving performance and efficiency. InstructCIR \cite{zhong2024compositional} focuses on composed image retrieval, which achieves better instruction-following capability by two-stage contrastive learning while progressive freezing the modules in the MLLM.

\paragraph{Evaluation} As the number of MLLM-based embedders increased, evaluation benchmarks with different focuses were proposed, such as M-BEIR \cite{wei2023uniir} for multi-modal retrieval, MVRB \cite{liu2025any} for multi-modal document understanding, MMTB \cite{jiang2024vlm2vec} for multi-modal universal embedding and UMRB \cite{zhang2024gme} for both uni-modal and multi-modal universal embedding. However, we find that many works do not indicate whether their evaluation is a true zero-shot out-of-domain or not, and there was a risk of unfair performance comparison. This is the reason why we do not list “evaluation-setting” in Table \ref{tab:mllm_embedders} as we do in Table \ref{tab:embedders}. We encourage subsequent work to filter the training data strictly to prevent overlapping training and evaluation data, thereby providing a fair performance benchmark for the community.

\subsection{Task-Specific Text Embedding}
For a long time, text embedding learning has been expected to obtain a universal embedding without revealing the downstream task. However, this setting is often difficult in practice, such as (1) In the unsupervised setting, contrastive learning brings the two views of data augmentation for the same instance closer together, i.e., learning an encoder that is insensitive to a particular data augmentation transform. However, this may result in valuable information for downstream tasks not being included in the embedding \cite{xiao2021should}; (2) In the supervised setting, undifferentiated joint training of different tasks will hurt performance \cite{wang2023relational,su2023one}; (3) In evaluation, the two texts can be evaluated from various perspectives and show different semantic similarities \cite{deshpande2023c,tu2024linguistically};

Therefore, we can convert the hope from universal text embeddings to universal text encoders $F$, which is expressed as
\begin{equation}
    h = F(x, t) \in \mathbb{R}^d.
\end{equation}

The universal encoders accept the text input $x$ and the downstream task $t$, and represent $x$ specifically for task $t$, where $t$ can be an identifier or a paragraph description of the task requirements. According to where $t$ acts, we can categorize the methods into pre-processing (also called instruction-following embedding) and post-processing (also called equivariant embedding learning).

\subsubsection{Instruction-Following Embedding} In instruction-following embedding learning, $F$ is divided into two parts: $F = f \circ p_t$, where $p_t$ is a prompt template related to task $t$. Initially, prompts are utilized to align the fine-tuning targets with the pre-trained targets of the PLMs \cite{zhong2021adapting}. Then, similar instruction tuning \cite{wei2022finetuned} is proposed to adjust LLMs to be more adaptable to the format of task-oriented conversations. Instruction-following embedding learning follows the idea of instruction tuning but models all tasks into a contrastive paradigm. TART \cite{asai2023task} firstly validates the feasibility of this idea for information retrieval, while Instructor \cite{su2023one} fine-tuned GTR on 330 NLP tasks and found that including instructions would alleviate conflicts between tasks compared to traditional multi-task learning. In the zero-shot evaluation, the model fine-tuned with instructions shows good generalization to new instructions and datasets. The findings related to this technique are still iterating: LLM2Vec \cite{behnamghader2024llm2vec} adds instruction only on evaluation and demonstrates its effectiveness; Inbedder \cite{peng2024answer} demonstrates that question-and-answer pairs can significantly improve a model's ability to comply with instructions under a particular training method.

\subsubsection{Equivariant Embedding Learning}
In equivariant Embedding learning, $F$ is divided into two parts: $F = u_t \circ f$, where $u_t$ is a mapping related to task $t$ in the embedding space. Before introducing specific methods, we define ``equivariant mapping'' in mathematics:

\begin{definition}\label{def:equivariant}
    {\bf (Equivariance Mapping)} Let $f: X \rightarrow \mathbb{R}^d$ is a mapping from data to embeddings. We call $f$ is equivariant to the algebraic group $T=(\mathbb{T}, \circ)$ if there exist a transformation $g: \mathbb{T} \times \mathbb{X} \rightarrow \mathbb{X}$ and a transformation $G: \mathbb{T} \times \mathbb{Z} \rightarrow \mathbb{Z}$ satisfying $G_t(f(x))=f(g_t(x))$ for any $t \in \mathbb{T}$ and $x \in \mathbb{X}$.
\end{definition}

$T$ is no longer restricted to algebraic groups in equivariant embedding learning. In practice, $t$ is viewed as a data augmentation strategy in the unsupervised setting or a mapping from data to data/label for a specific task in the supervised setting. Specifically, when $G_t$ is the identity mapping, the condition in Definition \ref{def:equivariant} degenerates to $f(x) = f(g_t(x))$, i.e., the alignment in contrastive learning \cite{wang2020understanding}. Instance-level equivariant embedding learning has been fully explored in self-supervised visual embedding learning \cite{dangovski2022equivariant, devillers2024equimod, suau2023duet, gupta2024structuring}. In NLP, equivariant embedding learning based on PLMs is applied in both task-level \cite{wang2023relational} and instance-level \cite{chuang2022diffcse,liu2023escl,yoo2024hyper}.

\subsection{Interpretable Text Embedding}
Current text embeddings are far superior at the performance level to those obtained using feature engineering; however, they severely lack interpretability. This leads to an inability to localize why they fail in some cases (e.g., bad cases in information retrieval) to the point where we can't target them for improvement. Some recent approaches attempt to enhance the interpretability of embeddings using LLMs.

\paragraph{Interpret Embedding with LLMs} \cite{tennenholtz2024demystifying} tries to use LLM interpretation tables to represent user portraits and movie information by arbitrary embeddings, even if these embeddings are constructed by human interpolation; \cite{teehan2024college} learn the embedding for the novel concept to explain them with LLMs;

\paragraph{Interpretable Embedding from LLMs} \cite{jiang2023scaling} add \texttt{The sentence [sent] means in a word:``} before the input text and using last pooling to obtain a text embedding. The method allows the text embedding to decode the token that summarizes the semantics of this text. Similarly, \cite{peng2024answer} trains the LLM using Q\&A pairs containing short answers and generates text embeddings using the last pooling strategy. \cite{nie2024text} find that the contrastive text embeddings produced by LLMs are highly correlated with the key tokens in the input text, independent of the model architecture, training strategy, and embedding method. \cite{benara2024crafting} directly ask LLMs predefined questions and generates text embeddings containing only 0-1 from the answers.

\section{Conclusion}
In this survey, we systematically explore the application of large language models (LLMs) to text embedding techniques, offering a unified perspective on the efforts made across various research communities for the first time. Our analysis reveals that the advent of LLMs has significantly mitigated challenges in text embedding, such as the scarcity of labeled data and limitations in model generalization. However, persistent issues, such as the difficulties posed by low-resource languages, remain unresolved. Moreover, new challenges, such as privacy concerns in text embedding, have emerged, necessitating further research and innovation in the future.

\section*{Contribution}
Zhijie Nie was responsible for the outline and versioning of this survey, writing Section 1, 2, 6, and 7, and involved in writing Section 4; Zhangchi Feng was responsible for formatting the dissertation of this survey, collecting the dissertation and writing for Section 5.1; Mingxin Li was responsible for collecting the dissertation and writing for Section 3; Cunwang Zhang was responsible for collecting the dissertation and writing for Section 5.2; Yanzhao Zhang was involved in writing Sections 2 and 4; and Dingkun Long was involved in writing Sections 4; Rizhong Zhang provided financial support and coordinated the review.

We are grateful for the efforts made by Hailang Huang for the advance preparation of this work. We thank Raghuveer Thirukovalluru for his suggestions for improving our paper.
\bibliographystyle{IEEEtran}
\bibliography{
    bibfiles/new,
    bibfiles/main,
    bibfiles/dataset,
    bibfiles/method_sts,
    bibfiles/method_ir,
    bibfiles/method_clustering,
    bibfiles/method_context_compression,
    bibfiles/method_universal,
    bibfiles/method_inversion
}

\end{document}